\documentclass[journal,twoside,web]{ieeecolor}
\usepackage{etoolbox} 
\makeatletter 
\@ifundefined{color@begingroup}%
{\let\color@begingroup\relax 
\let\color@endgroup\relax}{}%
\def\fix@ieeecolor@hbox#1{%
\hbox{\color@begingroup#1\color@endgroup}} 
\patchcmd\@makecaption{\hbox}{\fix@ieeecolor@hbox}{}{\FAILED} \patchcmd\@makecaption{\hbox}{\fix@ieeecolor@hbox}{}{\FAILED} 

\usepackage{tmi}
\usepackage{cite}
\usepackage{amsmath,amssymb,amsfonts}
\usepackage{algorithmic}

\usepackage{graphicx}
\usepackage{textcomp}
\usepackage{todonotes}
\usepackage{framed,multirow}
\usepackage{threeparttable}  
\usepackage{graphicx}
\usepackage{booktabs}
\usepackage[hidelinks]{hyperref}
\usepackage{amsmath}
\usepackage{float}
\usepackage{amssymb}
\usepackage{xcolor}
\usepackage{colortbl} 
\usepackage{paralist}
\usepackage{diagbox}
\usepackage{makecell}
\usepackage{soul}

\usepackage[capitalize]{cleveref}
\crefname{section}{Sec.}{Secs.}
\Crefname{section}{Section}{Sections}
\Crefname{table}{Table}{Tables}
\crefname{table}{Tab.}{Tabs.}

\def\BibTeX{{\rm B\kern-.05em{\sc i\kern-.025em b}\kern-.08em
    T\kern-.1667em\lower.7ex\hbox{E}\kern-.125emX}}
\begin{document}
\title{PET Head Motion {Estimation} Using Supervised Deep Learning with Attention}
\author{Zhuotong~Cai, \IEEEmembership{Student Member, IEEE}, Tianyi~Zeng, \IEEEmembership{Member, IEEE}, Jiazhen~Zhang, El\'{e}onore~V.~Lieffrig, Kathryn~Fontaine, Chenyu~You, \IEEEmembership{Member, IEEE}, Enette~Mae~Revilla, James~S.~Duncan, \IEEEmembership{Life Fellow, IEEE}, \\ Jingmin~Xin, \IEEEmembership{Senior Member, IEEE}, Yihuan~Lu, John~A.~Onofrey, \IEEEmembership{Member, IEEE}
\vspace{-2.5\baselineskip}
\thanks{©~2025 IEEE. This accepted manuscript has been accepted for publication in IEEE Transactions on Medical Imaging. 
DOI: 10.1109/TMI.2025.3620714. 
Personal use is permitted, but republication/redistribution requires IEEE permission.}
\thanks{
This work was supported by the National Institutes of Health (NIH) R21 EB028954.
(Zhuotong Cai and Tianyi Zeng contributed equally to this work.)
(Corresponding author: Tianyi Zeng; John Onofrey.)}
\thanks{Zhuotong Cai is with the National Key Laboratory
of Human-Machine Hybrid Augmented Intelligence, Xi'an Jiaotong University, Xi'an, Shannxi, China and Department of Radiology and Biomedical Imaging, Yale University, New Haven, CT 06519 USA. (email: cai199624@stu.xjtu.edu.cn; zhuotong.cai@yale.edu).}
\thanks{Tianyi~Zeng and Kathryn~Fontaine are with the Department of Radiology and Biomedical Imaging, Yale University, New Haven, CT 06519 USA. (e-mail: tianyi.zeng@yale.edu; kathryn.fontaine@yale.edu).}
\thanks{Jiazhen~Zhang and El\'{e}onore~V.~Lieffrig are with the Department of Biomedical Engineering, Yale University, New Haven, CT 06519 USA. (e-mail: jiazhen.zhang@yale.edu; eleonore.lieffrig@yale.edu).}
\thanks{Chenyu~You is with the Department of Electrical and Computer Engineering, Yale University, New Haven, CT 06511 USA. (e-mail: chenyu.you@yale.edu).}
\thanks{James~S.~Duncan is with the Department of Radiology and Biomedical Imaging, Department of Biomedical Engineering and Department of Electrical and Computer Engineering, Yale University, New Haven, CT 06519 USA. (e-mail: james.duncan@yale.edu).}
\thanks{Jingmin~Xin is with the National Key Laboratory
of Human-Machine Hybrid Augmented Intelligence, Xi'an Jiaotong University, Xi'an, Shannxi, 710049 China. (e-mail: jxin@mail.xjtu.edu.cn).}
\thanks{Enette~Mae~Revilla and Yihuan Lu are with United Imaging Healthcare, Shanghai, China. (e-mail: enette.revilla@united-imaging.com; yihuan.lu@united-imaging.com).}
\thanks{John~A.~Onofrey is with the Department of Radiology and Biomedical Imaging, Department of Biomedical Engineering and Department of Urology, Yale University, New Haven, CT 06519 USA. (e-mail: john.onofrey@yale.edu).}
\vspace{1em}
}
\maketitle

\begin{abstract}
Head movement poses a significant challenge in brain positron emission tomography (PET) imaging, resulting in image artifacts and tracer uptake quantification inaccuracies. Effective head motion estimation and correction are crucial for precise quantitative image analysis and accurate diagnosis of neurological disorders.
Hardware-based motion tracking (HMT) has limited applicability in real-world clinical practice.
To overcome this limitation, we propose a deep-learning head motion correction approach with cross-attention (DL-HMC++) to predict rigid head motion from one-second 3D PET raw data.
DL-HMC++ is trained in a supervised manner by leveraging existing dynamic PET scans with gold-standard motion measurements from external HMT.
We evaluate DL-HMC++ on two PET scanners (HRRT and mCT) and four radiotracers ($^{18}$F-FDG, $^{18}$F-FPEB, $^{11}$C-UCB-J, and $^{11}$C-LSN3172176) to demonstrate the effectiveness and generalization of the approach in large cohort PET studies.
Quantitative and qualitative results demonstrate that DL-HMC++ consistently outperforms {state-of-the-art data-driven motion estimation methods}, producing motion-free images with clear delineation of brain structures and reduced motion artifacts that are indistinguishable from gold-standard HMT.
Brain region of interest standard uptake value analysis exhibits average difference ratios between DL-HMC++ and gold-standard HMT to be 1.2$\pm$0.5\% for HRRT and 0.5$\pm$0.2\% for mCT.
DL-HMC++ demonstrates the potential for data-driven PET head motion correction to remove the burden of HMT, making motion correction accessible to clinical populations beyond research settings. The code is available at https://github.com/maxxxxxxcai/DL-HMC-TMI.
\end{abstract} 


\section{Introduction}
\label{sec:introduction}

Positron emission tomography (PET) imaging has gained prominence in human brain studies due to the availability of a diverse range of radiotracers. These radiotracers enable investigation of various neurotransmitters and receptor dynamics in different brain targets~\cite{nabulsi2016synthesis}, as well as studies of physiological or pathological processes~\cite{scholl2016pet}.
PET is commonly employed for diagnosis and monitoring of neurodegenerative diseases, including Alzheimer’s disease, Parkinson’s disease, epilepsy, and certain brain tumors~\cite{toyonaga2019vivo,sarikaya2015pet}. 
However, the presence of patient movement during PET brain scanning poses a significant obstacle to high-quality PET image reconstruction and subsequent quantitative analysis~\cite{keller2012methods}.
Even minor instances of head motion can substantially impact brain PET quantification, resulting in diminished image clarity, reduced concentrations in regions with high tracer uptake, and mis-estimation in tracer kinetic modeling~\cite{keller2012methods}. 
This problem is further exacerbated by the long duration of PET studies, where patients can involuntarily move~\cite {beyer2005use}. 
Even with physical head restraints, typical translations in the range of 5 to 20 mm and rotations of 1 to 4\textdegree are observed~\cite{rahmim2007strategies}.
Therefore, accurate monitoring and correction of head motion are critical for brain PET studies. 

{PET head motion estimation involves tracking patient movement during image acquisition, while motion correction (MC) refers to the process of compensating for the effects of head movement}~\cite{picard1997motion}.
Generally, patient movements in brain imaging are assumed to be of a rigid nature, composed of translation and rotation in three dimensions. 
The initial process to correct head motion involves motion estimation.
Once the motion information has been estimated, the motion-corrected PET image can be reconstructed using standard techniques such as frame-based or event-by-event (EBE) MC~\cite{jin2013evaluation}.
Therefore, accurate motion estimation is crucial for realizing high-quality PET imaging.

{Physical restraint during PET scanning can substantially reduce head motion effects. However, such methods cannot eliminate movement entirely, and this restrictive approach may be uncomfortable, especially over long scan durations, which reduces their acceptability for real-world use}~\cite{montgomery2006correction}.
Currently, head motion estimation methods are primarily categorized into the following types: 
\begin{inparaenum}[(i)]
\item hardware-based motion tracking (HMT), and
\item data-driven approaches.    
\end{inparaenum}
For HMT, high-frequency head motion information is provided by external devices.
Marker-based HMT, such as Polaris Vicra (NDI, Canada), tracks light-reflecting markers on the patient's head~\cite{jin2013evaluation}.
Despite its potential benefits, Vicra is not commonly employed in clinical practice because it necessitates the attachment of the marker to the patient. 
{Any inadvertent slippage or wobbling of the Vicra tool can introduce inaccuracies into the motion tracking process, thereby compromising the integrity of the data collected}~\cite{sun2022objective}.

Markerless HMT has also been developed for PET head motion estimation. 
Iwao et al.~\cite{iwao2022marker} applied a time-of-flight (TOF) range sensor to achieve markerless head motion tracking in a helmet PET system.
Slipsager et al.~\cite{slipsager2019markerless} and Zeng et al.~\cite{zeng2023markerless} applied camera systems in brain PET scans to achieve accurate high-frequency motion estimation. {However, these systems can be challenged by facial expressions and other non-rigid motions}~\cite{zhang2024data}. 
In general, HMT methods mainly rely on extra hardware support and setup, which limits their practical application in real-world clinical scenarios.

On the other hand, data-driven methods estimate head motion from reconstructions or PET raw data.
Spangler-Bickell et al.~\cite{spangler2022evaluation} utilized ultra-fast reconstruction methods to achieve motion estimation from short reconstruction frames in high-sensitivity and temporal resolution PET systems.
Revilla et al.~\cite{revilla2022adaptive} developed a data-driven head motion detection method based on the centroid of distribution (COD) of 3D PET cloud images (PCIs).
These methods utilized intensity-based image registration methods to align different frames, but these methods are sensitive to tracer kinetics and require manual parameter tuning. 
In contrast, deep learning (DL) methods, leveraging neural networks to construct a hierarchical representation of data through multiple layers of hidden units~\cite{lecun2015deep}, enable registration approaches to extract pertinent features directly from the data~\cite{fu2020deep}.
Salehi et al.~\cite{salehi2018real} proposed a DL model for medical image rigid registration and achieved real-time pose estimation of MRI.
Unsupervised DL methods were also developed for non-rigid medical image registration~\cite{de2019deep}.
Inspired by DL-based registration methods, Zeng et al.~\cite{zeng2022supervised} proposed a supervised DL head motion correction (DL-HMC) framework to predict rigid head motion information from PCIs using Vicra HMT as gold-standard motion information.
However, due to the noisy PCIs and limited generalization across data distributions, the effectiveness of these methods diminishes when applied to testing subjects that differ from the training dataset, especially when addressing subjects with significant movements.
{Subsequent DL methods have explored various strategies for PET head motion estimation.
Sundar et al. utilized conditional generative adversarial networks to synthesize pseudo high-count images from low-count PET brain images and applied frame-based registration for MC} \cite{sundar2021conditional}{, which ameliorated motion blurring to determine accurate motion information in an $^{18}$F-FDG study.
However, intra-frame motion can not be solved by frame-based MC, and the MRI navigators used in this study are challenging to implement with brain-dedicated PET scanners.
Lieffrig et al. }\cite{Lieffrig2023-nw} {developed a multi-task architecture for head MC, in which the rigid motion and motion-free PCI were predicted by the network.
The multi-task network enabled the model to learn the embedding of PCI representation, however, this network was sensitive to noise that introduced bias in testing subjects.
Reimers et al.}\cite{reimers2023deep}{ utilized a DL method to transform low-count images to high-count images, thereby predicting motion from high-quality subframes. However, training the network requires motion-free PET data, which is not available in this case.}

To address the limitations of the original DL-HMC approach, this study introduces an enhanced model, DL-HMC++, that incorporates a cross-attention mechanism, aiming to enhance {motion estimation and generalization performance}~\cite{Cai2023-ic}. 
Notably, attention mechanisms have demonstrated effective MC performance in cardiac image analysis applications~\cite{Ahn2023-dk,chen2022dual}. 
Our cross-attention mechanism takes a pair of features as input and computes their correlations to establish spatial correspondence between reference and moving PCIs. 
This explicitly enables the model to concentrate on the head region, which is the most relevant anatomy for motion estimation in brain PET studies.
This manuscript extends our previous work~\cite{Cai2023-ic} by
\begin{inparaenum}[(i)]
    \item including a rigorous validation of DL-HMC++ using a large cohort of human PET studies, {encompassing over 280 brain scans with 4 different tracers},
    \item providing extensive model analysis to assess generalization using two different PET scanners {with distinct TOF characteristics} and different tracers, {including cross-tracer generalization experiments}, 
    \item ablation studies to justify model design choices,
    \item quantitative evaluation of MC accuracy, and
    \item {comprehensive validation studies against several state-of-the-art (SOTA) benchmark motion estimation methods.}
\end{inparaenum}
Quantitative and qualitative evaluations demonstrate the robustness of DL-HMC++ across extensive experiments and highlight its ability to
correct head motion in PET studies {using only raw image data without the need for either reconstruction techniques or} HMT.

\section{Methods}
\label{sec:methods}

\subsection{Data-Driven Brain PET Motion Estimation Framework}

Our deep learning approach to brain PET head motion correction estimates rigid motion at one-second time resolution.
This data-driven motion estimation model utilizes one-second 3D PET cloud image (PCI) representations as input.
The reference $I_\text{ref}$ PCI and moving $I_\text{mov}$ PCI are created by back-projecting the PET listmode data from one-second time windows at times $t_\text{ref}$ and $t_\text{mov}$, respectively, along the line-of-response (LOR) with normalization for scanner sensitivity. 
For model training and evaluation, each one-second PCI has corresponding Vicra HMT information (rigid transformation matrix) as the gold-standard motion.
We train the model to estimate the rigid motion transformation $\theta = [t_x, t_y, t_z, r_x, r_y, r_z]$ between $I_\text{ref}$ and $I_\text{mov}$ where $\theta$ includes three translation ($t_d$) and three rotation ($r_d$) parameters for each axis $d = \{x,y,z\}$.

\begin{figure*}[t]
    \centering
    \includegraphics[width=0.9\textwidth]{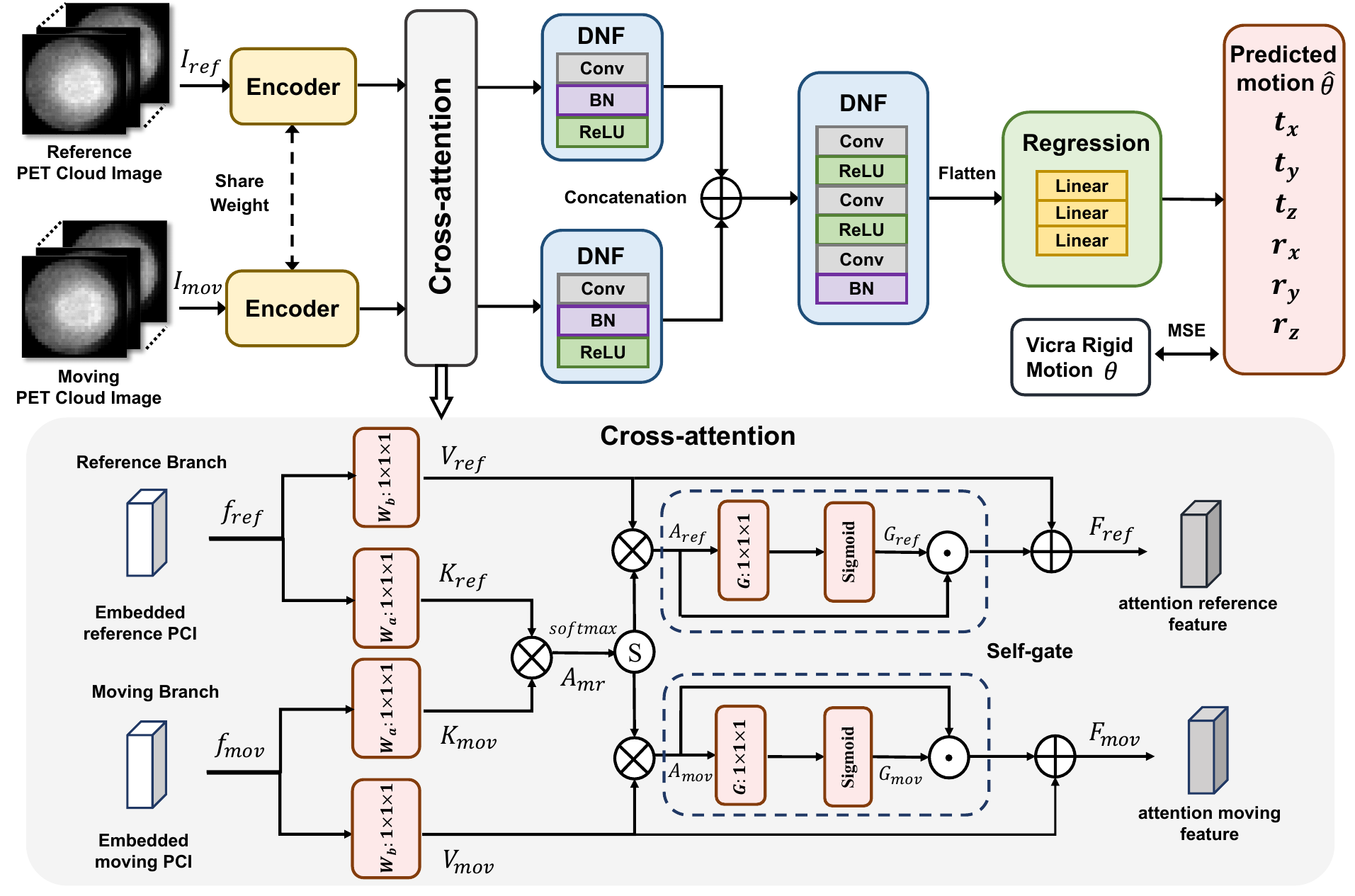}
    \caption{{\textbf{DL-HMC++ network architecture.} (Top) A shared encoder extracts imaging features from a pair of moving and reference PET cloud images. Then, the extracted features are fed into the cross-attention module to learn the correlation of anatomical features. 
    Deep Normalization and Fusion (DNF) blocks refine the attention features both before and after concatenation.
    Finally, concatenated attention features are fed into a multi-layer perceptron Regression block to predict motion. (Bottom) Details of the cross-attention module.
    }}
    \label{fig:architecture}
\end{figure*}

\subsection{DL-HMC++ Overview}

DL-HMC++ utilizes a supervised deep learning framework to predict rigid head motion for PET imaging. 
This network consists of three main components (\cref{fig:architecture}):  
\begin{inparaenum}[(i)]
    \item the feature extractor; 
    \item the cross-attention module; and,
    \item the regression layers. 
\end{inparaenum}
The feature extractor employs a shared-weight convolutional encoder to capture regional features from both the reference $I_\text{ref}$ and moving $I_\text{mov}$ PCIs. 
In contrast to our previous DL-HMC approach~\cite{zeng2022supervised} that used a ResNet encoder, here, we adopt a U-Net encoder~\cite{ronneberger2015u} with fewer parameters to extract features. 
Specifically, this encoder utilizes three convolutional layers with a 5$^3$  kernel size followed by a convolution with a 1$^3$  kernel, with the number of feature channels set to 32, 64, and 128, respectively.
We introduce a novel cross-attention mechanism to capture local features and global correspondence between the reference and moving PCIs, which will be elaborated in the following section. 
To enhance the representation of aggregated information following the cross-attention phase, we integrate a Deep Normalization and Fusion (DNF) module both prior to and after the concatenation process~\cite{mu2023learning}. 
The DNF module includes a series of convolutional layers, batch normalization and ReLU activation to refine the feature integration process.
Finally, a fully connected multi-layer perceptron (MLP) block takes the output of the final DNF block to infer the six rigid transformation motion parameters $\hat{\theta}$.

\begin{table*}[t]
\centering
\caption{{\textbf{PET Study Cohort.} The HRRT and mCT scanner cohorts are described in terms of sex, health status, injected activity, and motion information. Reported values are Mean$\pm$SD across subjects. In cohorts with a number of subjects greater than twenty, motion was computed on 20 randomly selected subjects to represent motion across the whole dataset.}}

\begin{tabular*}{\textwidth}{lcc|cc|cc|cc}
\toprule
& \multicolumn{4}{c}{HRRT}  & \multicolumn{4}{c}{mCT} \\
\cmidrule(lr){2-5} \cmidrule(lr){6-9} 
& \multicolumn{2}{c}{$^{18}$F-FDG}          & \multicolumn{2}{c}{{$^{11}$C-UCB-J}}        & \multicolumn{2}{c}{$^{18}$F-FPEB} & \multicolumn{2}{c}{{$^{11}$C-LSN3172176}} \\ 
\cmidrule(lr){2-3} \cmidrule(lr){4-5} \cmidrule(lr){6-7} \cmidrule(lr){8-9} 
& \multicolumn{1}{c}{Train} & Test & \multicolumn{1}{c}{{Train}} & {Test} & Train       & Test       & {Train}       & {Test}      \\ 
\midrule
N Subj. (M/F)   & 100 (56/44) & 20 (13/7) & {100 (53/45)} & {20 (16/4)} & 20 (8/12) & 4 (1/3) & {16 (7/9)} & {4 (4/0)} \\
Healthy Control & 42 & 7 & {37} & {8} & 20 & 4 & {8} & {2} \\
Alzheimer's Disease & 24 & 3 & {20} & {2} & 0 & 0 & {3} & {1} \\
Mild Cognitive Impairment & 19 & 1 & {9} & {0} & 0 & 0 & {5} & {1} \\
Epilepsy & 8 & 2 & {3} & {2} & 0 & 0 & {0} & {0} \\
Other & 7 & 7 & 31 & 8 & 0 & 0 & 0 & 0 \\
\midrule
Injected activity (mCi) & 4.83$\pm$0.28 & 4.93$\pm$0.15 & {14.99$\pm$5.15} & {14.91$\pm$4.84} & 3.75$\pm$1.19 & 4.47$\pm$0.16 & {14.27$\pm$4.43} & {15.77$\pm$6.32} \\ \midrule
Motion (mm) & 7.69$\pm$6.80 & 11.20$\pm$3.53 & {8.56$\pm$6.87} & {10.79$\pm$8.29} & 11.01$\pm$11.64 & 3.90$\pm$1.48 & {8.96$\pm$7.54} & {9.46$\pm$3.71}\\ \bottomrule
\end{tabular*}
\label{tab:data_description_table}
\end{table*}

\subsection{DL-HMC++ Cross-Attention}
Because of the ultra-short time duration (one-second), low system sensitivity, and lack of essential physical correction, {low-frequency bias} within the PCI significantly affects MC performance, making it challenging for the model to track head motion.
To mitigate the impact of noise and to enhance motion estimation performance, we introduce the attention mechanism in our model to emphasize the head region.  
This module establishes spatial correspondences between features derived from the reference image and those from the moving image.
It takes two inputs $f_\text{ref}\in \mathbb{R}^{C \times H \times W \times D}$ and $f_\text{mov}\in \mathbb{R}^{C \times H \times W \times D}$, which represent the feature maps of the reference and moving images, respectively, where
$H, W$ and $D$ denote the feature map dimensions and $C$ denotes the number of feature channel.
Initially, we partition $f_\text{ref}$ into reference key $K_\text{ref}$ and value $V_\text{ref}$ and, likewise, $f_\text{mov}$ is divided into moving query $K_\text{mov}$ and value $V_\text{mov}$:
\begin{align*}
K_\text{ref} & = W_{a}  f_\text{ref}, \quad V_\text{ref} = W_{b}   f_\text{ref}, \\
K_\text{mov} & = W_{a}  f_\text{mov}, \quad V_\text{mov} = W_{b}  f_\text{mov},
\end{align*}
where $W_a, W_b$ are the $1\times1\times1$ convolution layers.
We reshape $K_\text{mov}$ and $K_\text{ref}$ to the dimension of $C \times (HWD)$ and calculate the attention matrix using the following equation:
\begin{equation*}
A_\text{mr} = \text{Softmax}(K_\text{mov}^T K_\text{ref}) \in R^{(HWD)\times(HWD)},
\end{equation*}
where $A_\text{mr}$ represents the similarity matrix, correlating each row of  $K_\text{mov}^T$ with each column of $K_\text{ref}$. Upon computing the attention map $A_{mr}$, the attention features are updated for both the reference and moving features as follows:
\begin{equation*}
    A_\text{ref} = A_\text{mr} \cdot V_\text{ref} , \quad
    A_\text{mov} = A_\text{mr}^T \cdot V_\text{mov}.
\end{equation*}

To enhance the model's ability to identify and prioritize the most critical feature representation for the motion analysis between the moving and reference PCIs, we incorporate a self-gating mechanism. This approach assigns variable weights to the input data, enabling the model to discern and selectively integrate relevant information from both the moving and reference PCIs. The gating mechanism is operated by dynamically adjusting the contribution of each input, ensuring that the most informative parts have a greater influence on the outcome of the motion estimation, which is formulated as follows:
\begin{equation*}
G_\text{ref}, G_\text{mov} = \sigma(G(A_\text{ref})), \sigma(G(A_\text{mov}))  \in [0,1]^{HWD},
\end{equation*}
where $\sigma$ is the logistic sigmoid activation function and {$G$ is the   1$\times$1$\times$1 convolution layer.} 
The gate module effectively determines the extent to which information from the PCI could be retained and automatically learned during the training. By applying this gate across the features of both the moving and reference PCIs, the model generates a weighted combination that emphasizes the most relevant features for motion analysis. 
This results in an enriched feature representation that captures the essential details from both images, facilitating a more precise and informed estimation of motion. The final attention feature representations for both the moving and reference features are derived as follows:
\begin{align*}
    F_\text{ref} = G_\text{ref} A_\text{ref} + V_\text{ref},\quad
    F_\text{mov} = G_\text{mov} A_\text{mov} + V_\text{mov}.
\end{align*}

\section{Results}

{We validate DL-HMC++'s effectiveness for head motion estimation using a diverse set of brain PET studies from two different scanners.
We compare performance with multiple motion estimation baselines and provide ablation studies to justify model design choices. 
Finally, we demonstrate accurate motion estimation and correction through rigorous quantitative and qualitative evaluations.}

\subsection{Experimental Setup}

\subsubsection{Data}

We retrospectively identified a cohort of existing brain PET studies from the Yale PET Center.
The cohort contains a diverse set of PET data from {four different radiotracers} acquired on two different scanners:
\begin{inparaenum}[(i)]{
\item 120 $^{18}$F-FDG {and 120 $^{11}$C-UCB-J}~\cite{chen2021comparison} scans acquired on a brain-dedicated High Resolution Research Tomograph (HRRT) scanner (Siemens Healthineers, Germany) {without time-of-flight (TOF)}; and
\item 24 $^{18}$F-FPEB~\cite{lim2014preparation} {and 20 $^{11}$C-LSN3172176}~\cite{naganawa2021first} scans acquired on a conventional mCT scanner (Siemens Healthineers, Germany) {with TOF}.
}\end{inparaenum}
The datasets contain a diverse mix of subjects and clinical conditions that include healthy controls, neurological disorders such as Alzheimer's Disease (AD), mild cognitive impairment (MCI), epilepsy, and other diagnoses.
We divide each dataset into Training, Validation and Testing sets using an 8:1:1 ratio (\cref{tab:data_description_table}).
All scans include Vicra HMT information used as gold-standard motion estimation, T1-weighted magnetic resonance imaging (MRI), PET-space to MRI-space transformation matrices, and FreeSurfer anatomical MRI segmentations~\cite{fischl2012freesurfer}. 
All PET imaging data is 30 minutes acquired from 60-minutes post-injection.
{Summary estimates of head motion magnitude were quantified over the entire scan duration using the method described by Jin et al. in}~\cite{jin2013evaluation}.
All subjects were enrolled in studies approved by the Yale Institutional Review Board and Radiation Safety Committee with written informed consent.

\subsubsection{Evaluation Metrics}
{We evaluate head motion estimation performance using quantitative and qualitative assessment.}

\paragraph{Quantitative Assessment of Motion Estimation}
\label{sec:Quantitative_Assessment}
{To quantitatively evaluate the performance of motion estimation, we calculate the Root Mean Squared Error (RMSE) between the estimated motion parameters ($\hat{\theta}$) and the Vicra gold-standard ($\theta$). The RMSE was computed for each individual motion component (translation and rotation) separately across the full scan duration.} 
To robustly summarize motion estimation performance, we calculate the {mean value and standard deviation (SD) of the RMSE error across all testing subjects}.
We assess the statistical significance of DL-HMC++ compared to other MC methods on the HRRT dataset using a two-tailed Wilcoxon signed-rank test to evaluate if the DL-HMC++ {RMSE} result is smaller than that of the other methods. 
{The Wilcoxon signed-rank test was selectively applied to the HRRT's 
$^{18}$F-FDG and $^{11}$C-UCB-J datasets, but did not apply to the mCT datasets due to the test set sample size (n$=$4 subjects) being below the minimum requirement (n$\geq$6).}

\paragraph{Qualitative and Quantitative Assessment of Reconstructed PET Images}
\label{sec:method:roi}

For HRRT $^{18}$F-FDG and mCT $^{18}$F-FPEB studies, we qualitatively compare MOLAR reconstructed images by visual inspection {and quantitatively assess differences by computing normalized error maps $E_\text{pred}$. 
Here,
$E_\text{pred} = (R_\text{pred}-R_\text{Vicra})/ \max{(|R_\text{pred}-R_\text{Vicra}|)}$ (scale to the range [$-1$, $1$]), where $R_{pred}$ and $R_\text{Vicra}$ are the reconstructed images from motion-correction and Vicra HMT, respectively.}
To evaluate the final motion-corrected PET reconstruction images quantitatively, we perform brain ROI analyses using the FreeSurfer~\cite{fischl2012freesurfer} segmented ROI masks to quantify mean standard uptake value (SUV) within each ROI. 
We aggregate the original 109 FreeSurfer ROIs into 14 grey matter (GM) ROIs: Amygdala; Caudate; Cerebellum Cortex; Frontal; Hippocampus; Insula; Occipital; Pallidum; Parietal; Putamen; Temporal; Thalamus; and two white matter (WM) ROIs, the Cerebellum and Cerebral WM.
We perform a bias-variance analysis between the mean SUV within each ROI and the SUV derived using the Vicra gold-standard by computing the absolute difference ratio.

To evaluate performance at anatomically meaningful locations, we calculate the mean distance error (MDE) of anatomical brain ROIs~\cite{revilla2022adaptive}.
Using the FreeSurfer segmented ROI masks, we calculate the center-of-mass (COM) for each ROI on the Vicra MC result (COM$_{\text{Vicra}}$).
Then, the same ROI masking is applied to the MOLAR reconstruction images with different MC methods, and the estimated COM (COM$_{\text{est}}$) of each method is calculated. 
The MDE is defined as the mean of the Euclidean distance between COM$_{\text{Vicra}}$ and COM$_{\text{est}}$ across all ROIs. 
A larger MDE indicates worse motion estimation.

\subsubsection{Cross-tracer Generalization Evaluation}
\label{sec:method:cross-tracer}
{To validate the model's cross-tracer generalization capability, we conduct a comprehensive evaluation by directly applying the model weights trained on $^{11}$C datasets to perform inference on $^{18}$F datasets without any fine-tuning or parameter adjustment. Specifically, the model weights obtained from HRRT $^{11}$C-UCB-J training are applied to $^{18}$F-FDG data, while the weights from mCT $^{11}$C-LSN3172176 training are evaluated on $^{18}$F-FPEB data. 
Quantitative assessment of motion estimation is conducted by comparing the model's performance on these unseen tracers with the gold-standard Vicra, evaluating RMSE for both translation and rotation parameters} (Sec.~\ref{sec:Quantitative_Assessment}). {This evaluation provides critical insights into the model's robustness and generalizability across diverse tracer applications.}

\subsubsection{Baseline Motion Estimation Methods}
We comprehensively compared our approach for head motion {estimation against SOTA benchmark methods, including intensity-based registration and deep learning methods}: 
\begin{inparaenum}[(i)]
    \item BioImage Suite~\cite{papademetris2006bioimage} (BIS), an intensity-based rigid registration procedure that minimizes the sum-of-squared differences; 
    \item SimpleElastix~\cite{marstal2016simpleelastix} (SIM), a widely utilized medical image registration tool that employs mutual information as a similarity metric to rigidly register the PCIs; 
    \item Imregtform~\cite{muthukumaran2017medical} (IMR), a medical image registration method that uses intensity-based rigid registration algorithm with {MSE loss}, which was used in prior data-driven PET head MC studies~\cite{spangler2022evaluation};
    \item DL-HMC \cite{lieffrig2022multi}, our prior supervised deep learning approach for head MC that includes a time-conditioning module and excludes attention; 
    \item DL-HMC without time-conditioning (DL-HMC w/o TC), which removes the time conditional module from the original DL-HMC;
    and
    \item Dual-Channel Squeeze-Fusion-Excitation (DuSFE)~\cite{chen2023dusfe}, a deep learning registration approach designed to extract and fuse the input information for cross-modality rigid registration.
\end{inparaenum}
{To further enhance the registration quality of the intensity-based methods, following the same workflow in}~\cite{spangler2021optimizing}{, high-resolution one-second fast reconstruction images (FRIs) were generated using CPU-parallel reconstruction platforms for the mCT dataset}~\cite{zeng2023fast}{. We evaluated BIS and IMR using FRIs as inputs during the mCT experiments. 
No motion correction (NMC) results were also compared for reference.}

\subsubsection{Implementation Details}
\paragraph{Data Processing}

To create the DL-HMC++ input, we pre-process the HRRT PCI data volumes by downsampling from 256$\times$256$\times$207 voxels (1.22$\times$1.22$\times$1.23 mm$^3$) to 32$\times$32$\times$32 voxels (9.76$\times$9.76$\times$7.96 mm$^3$) using area interpolation. 
Similar pre-processing is applied to mCT PCI data from 150$\times$150$\times$109 voxels (2.04$\times$2.04$\times$2.03 mm$^3$ voxel spacing) to 32$\times$32$\times$32 voxels (9.56$\times$9.56$\times$6.91 mm$^3$ voxel spacing).

\paragraph{Network Training}

To efficiently train the network, we randomly sub-sample 360 out of 1,800 time points for each study in the training set. 
During each training epoch, we randomly pair two PCIs as reference $I_\text{ref}$ and moving $I_\text{mov}$ image inputs, such that $t_\text{mov}>t_\text{ref}$, and calculate their relative Vicra motion on the fly.
We train the network using a mini-batch size of 12 and minimize the mean squared error (MSE) between the predicted motion estimate $\hat{\theta}$ and Vicra $\theta$ using Adam optimization with initial learning rate 5e-4, $\gamma$=0.98, and exponential decay with step size 200 for training. 

\paragraph{Network Inference}
For inference on testing subjects independent of the training data, we utilize a single reference PCI $I_\text{ref}$ at the first time point and register all following PCIs at the remaining time points to estimate the rigid transformation to the reference space $I_\text{ref}$.

\paragraph{Event-by-Event (EBE) Motion Compensated Reconstruction}
Once the rigid motion transformation parameters have been estimated by DL-HMC++, we reconstruct the PET image using the EBE motion compensation OSEM list-mode algorithm for resolution-recovery reconstruction (MOLAR)~\cite{jin2014evaluation}. 
MOLAR reassigns the endpoints of each LOR according to the motion estimation result to reconstruct the motion-corrected PET image.
For HRRT studies, OSEM reconstruction (2 iterations $\times$ 30 subsets) with spatially invariant point-spread-function (PSF) of 2.5-mm full-width-half-maximum (FWHM) is applied with reconstruction voxel size 1.22$\times$1.22$\times$1.23 mm$^3$.
For mCT studies, OSEM reconstruction (3 iterations $\times$ 21 subsets) with spatially invariant PSF of 4.0-mm FWHM is applied with
reconstruction voxel size 2.04$\times$2.04$\times$2.00 mm$^3$.

\begin{figure*}[t]
    \includegraphics[width=\textwidth]{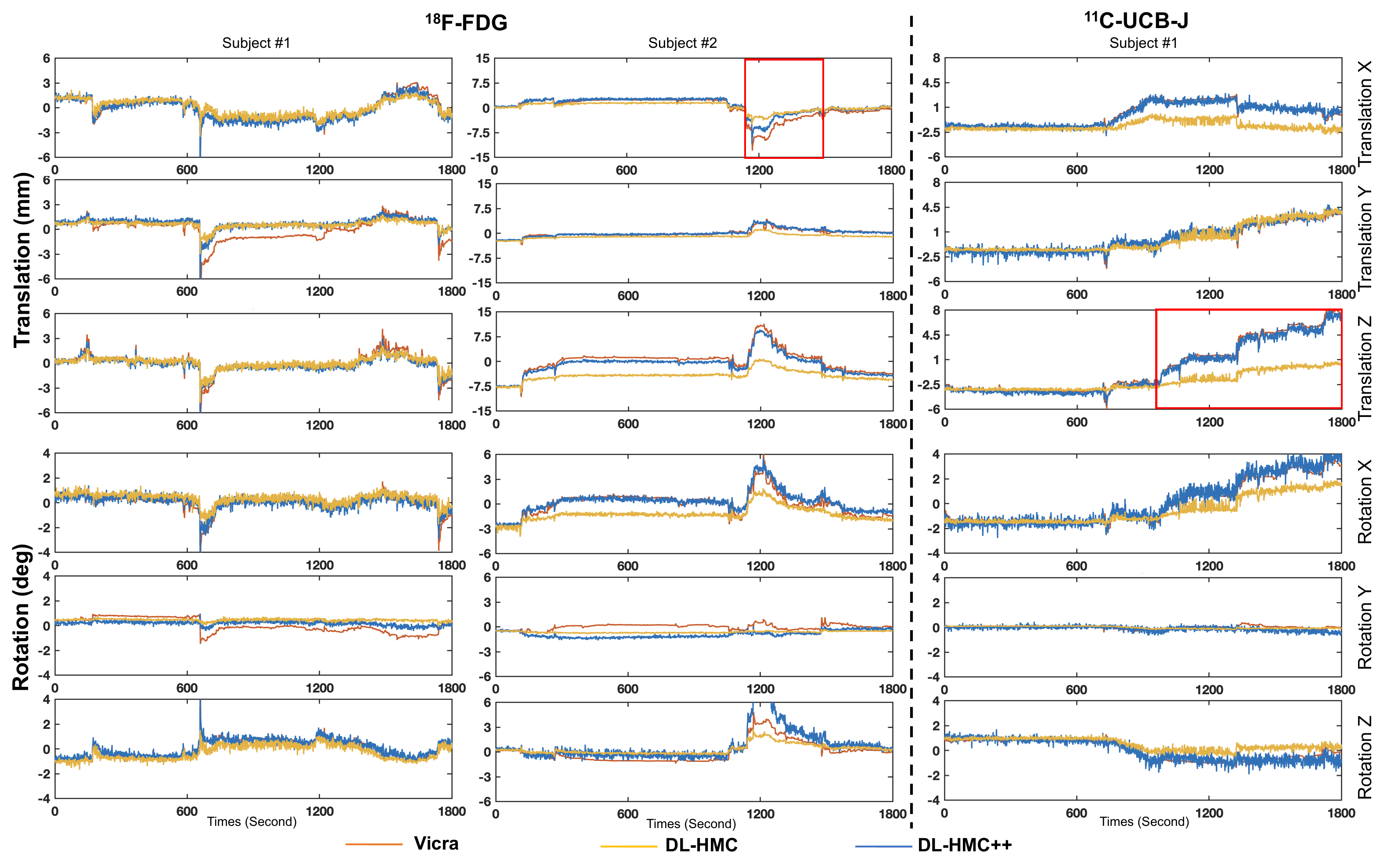}
    \vspace{-1.5\baselineskip}
    \caption{{\textbf{HRRT motion prediction results with $^{18}$F-FDG and $^{11}$C-UCB-J tracers}. 
    Each column represents a unique subject from the test set. 
    Rows show rigid transformation parameters
    (from top to bottom: translation in x, y, z directions and rotation about the x, y, z axes) }from gold-standard Vicra (\textcolor[RGB]{237,125,49}{orange}), DL-HMC (\textcolor[RGB]{255,192,0}{yellow}) and our proposed method (\textcolor[RGB]{68, 114, 196}{blue}) on HRRT. 
    Red boxes indicate time intervals of interest for DL-HMC++ performance.}
    \label{fig:hrrt_results}
\end{figure*}

\subsection{HRRT Results} 
\subsubsection{$^{18}$F-FDG} DL-HMC++ demonstrates the best quantitative motion estimation performance compared to all other benchmark methods, with translation and rotation {RMSE of 1.27 mm and 1.16\textdegree, respectively }(\Cref{tab:all_prediction}). 
{The Wilcoxon signed-rank test reveals that DL-HMC++ achieves statistically significant improvements (p\textless0.05) in both translation and rotation errors compared to all benchmark methods.}
Overall, DL methods outperform the intensity-based registration approaches with more accurate and effective motion estimation results. 
DL-HMC++ significantly outperformed original DL-HMC, demonstrating a {49\% and 27\%} improvement in translation and rotation, respectively.

\Cref{fig:hrrt_results} visualizes DL-HMC++ motion estimation results with respect to the original DL-HMC and the Vicra gold-standard, which demonstrates that the proposed method can effectively track head motion. {In FDG Subject \#1, both models demonstrate excellent alignment with actual Vicra head motion patterns. For Subject \#2, a poor performance occurs in translation X (red bounding box), where DL-HMC++ shows a misalignment with Vicra, however, DL-HMC exhibits larger errors.
This mismatch may be attributed to the substantial distance between the moving frame and the reference frame.
Moreover, our model performs well during other periods, demonstrating its capability to estimate movements with relatively large translations over 15 mm and 9-degree rotations.} 

\begin{figure}[!t]
    \centering
    \includegraphics[width=1.0\columnwidth]{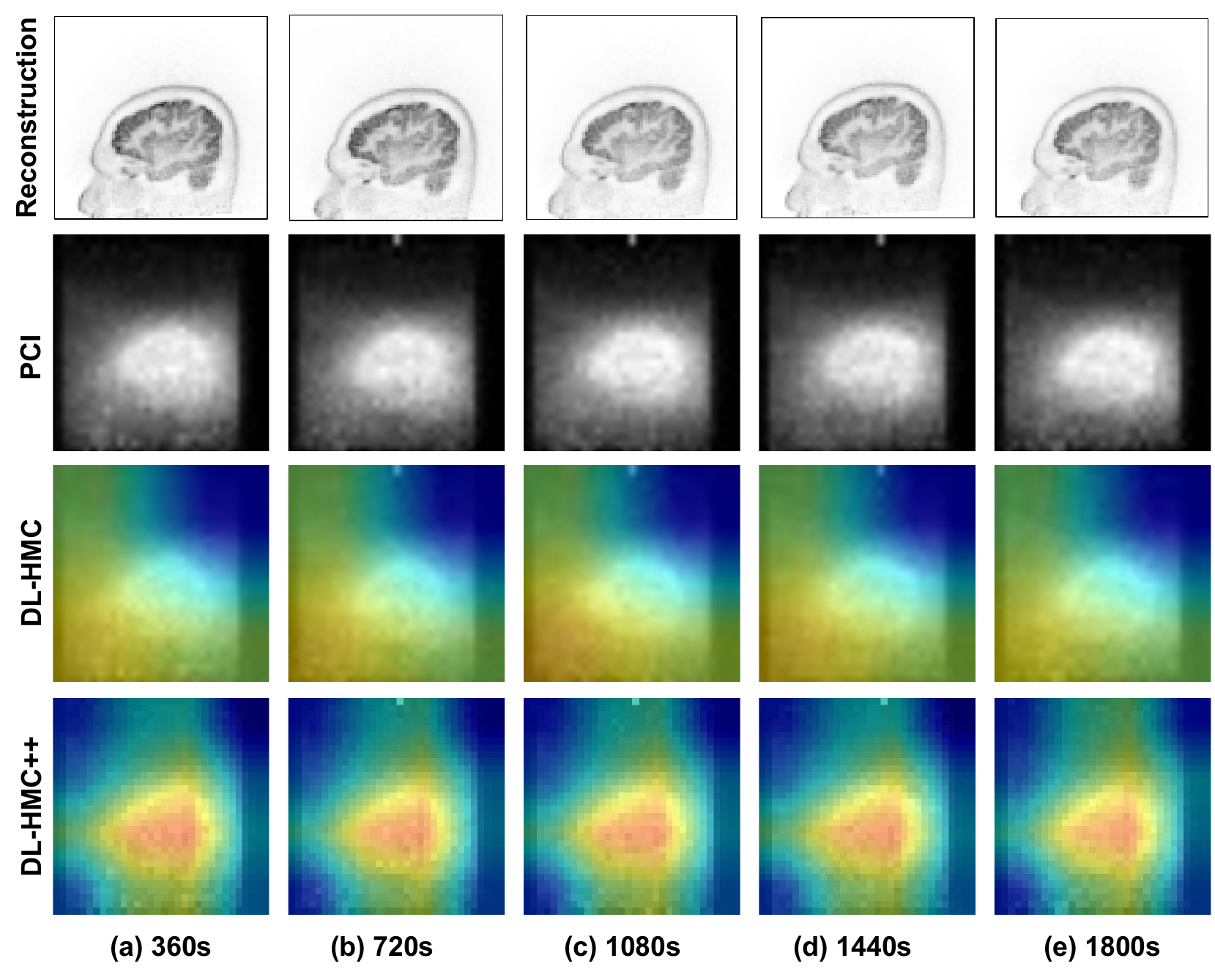}
    \vspace{-1.8\baselineskip}
    \caption{{\textbf{Grad-CAM saliency map visualization.} Sagittal view from five different time frames of the HRRT testing set during 30 min (1,800 s) PET acquisition. Our proposed DL-HMC++ method more accurately localizes the head anatomy compared to DL-HMC without attention.}}
    \label{fig:gradcam}
\end{figure}

In addition, DL-HMC++'s proposed cross-attention module enhances the model's ability to correct motion by concentrating on the head region during the motion tracking, which we confirm using Grad-CAM~\cite{selvaraju2017grad} to visualize saliency maps and compare to DL-HMC (\cref{fig:gradcam}). 
DL-HMC's saliency maps highlight areas outside the head, suggesting this model failed to focus on the relevant anatomical information in the PCI.

\begin{table*}[t]
\centering
\small
\caption{{\textbf{Quantitative motion estimation results.} Motion prediction RMSE error of translation (Trans.) (mm) and rotation (Rot.) (degrees) components compared to Vicra gold-standard on two PET scanners (HRRT and mCT) using four radiotracers ($^{18}$F-FDG, $^{18}$F-FPEB, $^{11}$C-UCB-J, and $^{11}$C-LSN3172176). Reported values are Mean$\pm$SD.}}
\label{tab:all_prediction}

\begin{tabular*}{\textwidth}{l@{\extracolsep{\fill}}cc|cc|cc|cc}

\toprule
\multirow{4}{*}{Method} &
\multicolumn{4}{c}{HRRT} & \multicolumn{4}{c}{mCT} \\
\cmidrule(lr){2-5}  \cmidrule(lr){6-9}
& \multicolumn{2}{c}{$^{18}$F-FDG} & \multicolumn{2}{c}{$^{11}$C-UCB-J} & \multicolumn{2}{c}{$^{18}$F-FPEB} & \multicolumn{2}{c}{$^{11}$C-LSN3172176} \\ 
\cmidrule(lr){2-3}  \cmidrule(lr){4-5} \cmidrule(lr){6-7} \cmidrule(lr){8-9}
& Trans. (mm) & Rot. (deg) & Trans. (mm) & Rot. (deg) & Trans. (mm) & Rot. (deg) & Trans. (mm) & Rot. (deg)  \\ 
\midrule
NMC & 6.29$\pm$5.79$^*$ & 3.12$\pm$1.42$^*$ & 6.86$\pm$19.58$^*$ & 3.27$\pm$6.14$^*$  & 2.42$\pm$1.43 & 1.36$\pm$0.48 & 4.63$\pm$7.76 & 2.10$\pm$1.36 \\ 
\midrule
BIS & 4.26$\pm$5.31$^*$ & 2.06$\pm$3.01$^*$ & 3.18$\pm$3.56$^*$ & 1.63$\pm$1.54$^*$  & 1.32$\pm$0.06 & 0.53$\pm$0.05 & 1.40$\pm$0.20 & 0.66$\pm$0.06 \\ 
SIM & 3.15$\pm$4.87$^*$ & 1.94$\pm$2.70$^*$ & 3.04$\pm$2.53$^*$ & 1.58$\pm$1.32$^*$
 & 1.57$\pm$0.10 & 1.24$\pm$0.02 & 3.06$\pm$2.05 & 2.60$\pm$3.03 \\
IMR & 2.84$\pm$3.83$^*$ & 2.25$\pm$2.85$^*$ & 3.52$\pm$3.97$^*$ & 1.77$\pm$1.50$^*$ 
 & 1.38$\pm$0.28 & 0.55$\pm$0.05 & 2.32$\pm$2.26 & 0.88$\pm$0.07 \\
\midrule 
DL-HMC & 2.49$\pm$2.43$^*$ & 1.59$\pm$2.32$^*$ & 2.07$\pm$1.87$^*$ & 1.35$\pm$1.09$^*$ & 0.93$\pm$0.20 & 0.40$\pm$0.03 & 1.46$\pm$0.35 & 0.71$\pm$0.09 \\
-w/o TC & 1.76$\pm$1.19$^*$ & 1.33$\pm$1.63$^*$ & 1.54$\pm$0.62$^*$ & 1.34$\pm$1.13$^*$ & 0.80$\pm$0.01 & 0.57$\pm$0.01 & 1.19$\pm$0.11 & 0.61$\pm$0.02 \\
DuSFE & 1.56$\pm$0.66$^*$ & 1.37$\pm$1.73$^*$ & 1.36$\pm$0.46~ & 1.36$\pm$0.85$^*$ & 0.60$\pm$0.03 & 0.41$\pm$0.02 & 1.21$\pm$0.12 & 0.69$\pm$0.10 \\
DL-HMC++ & \textbf{1.27$\pm$0.46~} & \textbf{1.16$\pm$1.20~} & \textbf{1.26$\pm$0.44~} & \textbf{1.22$\pm$0.98~} & \textbf{0.54$\pm$0.00} & \textbf{0.40$\pm$0.00} & \textbf{0.99$\pm$0.02} & \textbf{0.58$\pm$0.03} \\
\bottomrule
\end{tabular*}
\begin{tablenotes}
\item{Note:}  $^*$ indicates p $<$ 0.05.
\end{tablenotes}
\end{table*}

\subsubsection{$^{11}$C-UCB-J}

{The performance evaluation on $^{11}$C data from HRRT demonstrates consistent superiority of DL-HMC++, similar to its performance on $^{18}$F data} (\cref{tab:all_prediction}). 
{
Quantitative results indicate that DL-HMC++ achieves the best performance across all evaluation metrics,
with translation and rotation RMSE values of 1.26 mm and 1.22\textdegree, respectively.
Statistical evaluation confirms that DL-HMC++ achieves significantly superior performance over nearly all benchmark methods (p\textless0.05).  
Compared to the original DL-HMC, DL-HMC++ demonstrates a 39\% improvement in translation and a 10\% improvement in rotation.}

{Visualizing the motion prediction results for one $^{11}$C subject in HRRT} (\cref{fig:hrrt_results}, third column), 
{DL-HMC++ demonstrates promising capability in capturing large motion patterns, even under challenging conditions (e.g., 14 mm in z-axis translation and 7{\textdegree} in x-axis rotation).
Compared to the original DL-HMC, DL-HMC++ achieves superior motion detection sensitivity. For example, as highlighted by the red bounding box, DL-HMC++ benefits from the enhanced attention module to precisely predict both the motion trend and magnitude, even for a 10 mm movement.}

\begin{figure*}[t]
    \includegraphics[width=\textwidth]{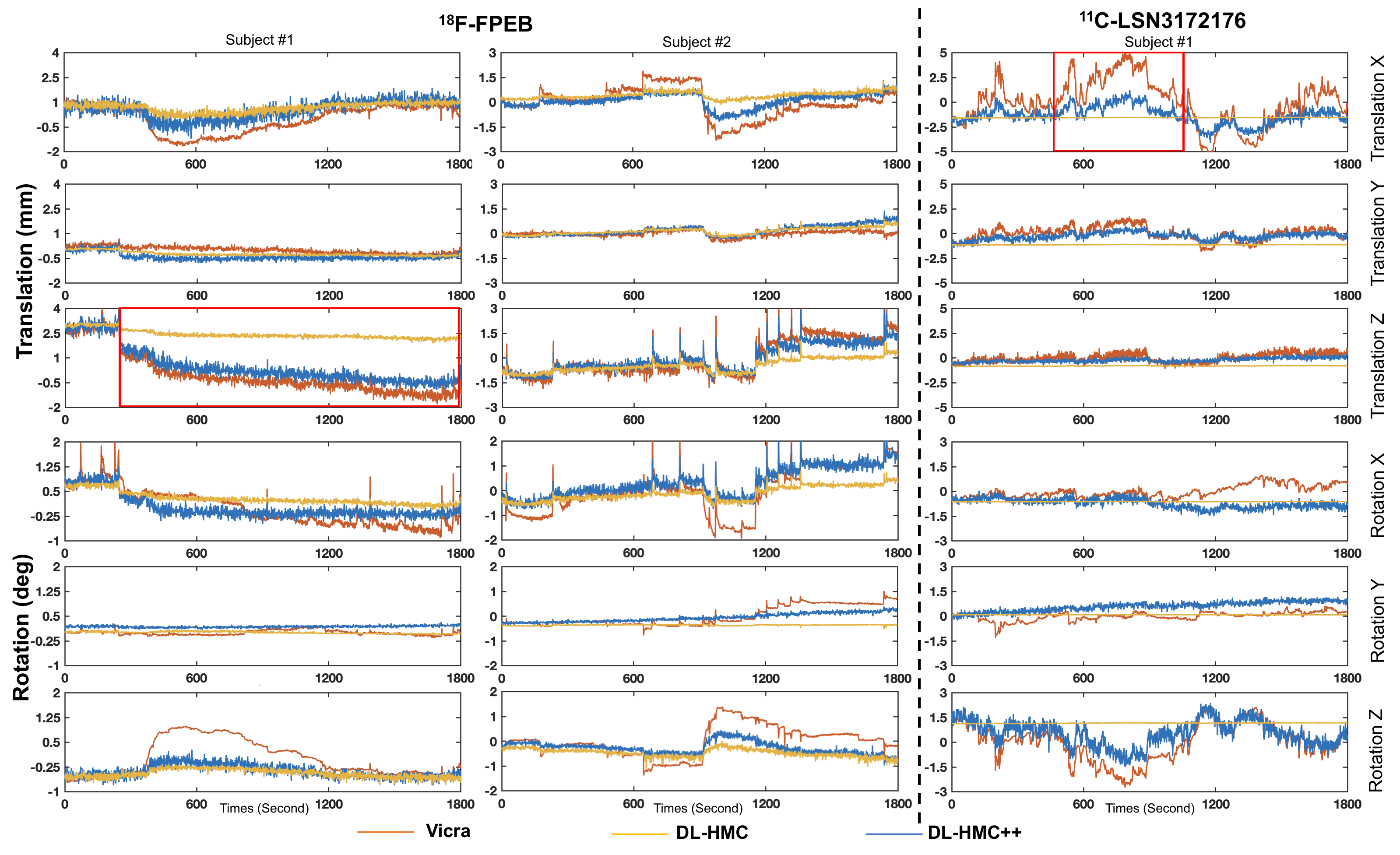}
    \vspace{-1.8\baselineskip}
    \caption{{\textbf{mCT motion prediction results with $^{18}$F-FPEB and $^{11}$C-LSN3172176 tracers}. Each column represents a unique subject from the test set. 
    Rows show rigid transformation parameters
    (from top to bottom: translation in x, y, z directions and rotation about the x, y, z axes)} from gold-standard Vicra (\textcolor[RGB]{237,125,49}{orange}), DL-HMC (\textcolor[RGB]{255,192,0}{yellow}) and our proposed method (\textcolor[RGB]{68, 114, 196}{blue}) on mCT. 
    Red boxes indicate time intervals of interest for DL-HMC++ performance.}
    \label{fig:mct_results}

\end{figure*}

\subsection{mCT Results}

\subsubsection{ $^{18}$F-FPEB}
DL-HMC++ remains competitive on the mCT $^{18}$F-FPEB data, reaching {RMSE of 0.54 mm in translation and 0.40{\textdegree} in rotation} (\Cref{tab:all_prediction}) on the testing dataset. 
We observe a consistent trend between intensity-based registration methods and DL methods from the HRRT to mCT, where DL methods outperform {SOTA image-intensity registration methods (BIS, IMR) that even utilize FRIs as input}~\cite{spangler2021optimizing,zeng2023fast}.
Similar to the HRRT results, DL-HMC++'s attention mechanism helps capture the motion with better estimation performance.
It is also noticeable that DL-HMC++ ranked the best in both translation and rotation error, outperforming the original DL-HMC by {42\%} in translation.

\Cref{fig:mct_results}{ shows the motion prediction results for the $^{18}$F-FPEB dataset, comparing DL-HMC++ with the baseline DL-HMC and the Vicra gold standard. 
While the overall performance on mCT data is less accurate than on HRRT data likely due to relatively fewer training data samples, DL-HMC++ demonstrates notable improvements over DL-HMC. A key example is in $^{18}$F-FPEB Subject \#1 (translation Z), where DL-HMC fails to track the motion (red bounding box), while DL-HMC++ successfully detects the substantial movements. 
In $^{18}$F-FPEB Subject \#2, both DL-HMC and DL-HMC++ underestimate rotations on the x-axis and z-axis, however, this error is limited to $\sim$1.5{\textdegree}.}

\subsubsection{ $^{11}$C-LSN3172176} 
{Building upon the promising results demonstrated with $^{18}$F in mCT, our proposed DL-HMC++ framework maintains superior performance in both translation and rotation estimation for the more challenging  $^{11}$C-LSN3172176. The quantitative results in} \Cref{tab:all_prediction} {reveal that DL-HMC++ outperforms all benchmark methods, demonstrating an 18\% improvement in translation and 16\% improvement in rotation compared to DuSFE. }

{The $^{11}$C subject \#1 visualization in} \Cref{fig:mct_results}  {further presents a noteworthy observation. While DL-HMC fails to capture motion information, as demonstrated by its flattened prediction curve, our proposed DL-HMC++ algorithm maintains robust performance. Although the red bounding box indicates an intensity mismatch with Vicra due to continuous movements with relatively large and rapid amplitudes, DL-HMC++ successfully detects the overall movement trends up to 10 mm in translation X and 4\textdegree ~in rotation Z.}

{In summary, the significant improvements in motion estimation achieved by DL-HMC++ over other methods across diverse scenarios and challenging conditions underscore the enhanced robustness of our proposed method.}

\begin{table}[t]
 \caption{
 \textbf{{Ablation Studies.}} {Motion prediction of translation and rotation transformation RMSE compared to Vicra gold-standard motion tracking on the HRRT $^{18}$F-FDG datasets for Network Architecture, Attention Type, Cloud Size and Subject Number. Reported values are Mean$\pm$SD.}
} 
\begin{center}
\begin{tabular*}{\columnwidth}{c @{\extracolsep{\fill}}ccc}
\toprule
\multicolumn{2}{c}{Ablation Part} & Trans. (mm) & Rot. (deg) \\ 
\midrule
\multirow{4}{*}{\makecell[c]{Network\\Arch.}} & Proposed & \textbf{1.27$\pm$0.46}  & \textbf{1.16$\pm$1.20}  \\
& w/o gate &  1.52$\pm$0.52 & 1.37$\pm$1.98 \\  
& w/o DNF &1.62$\pm$1.03 & 1.33$\pm$1.77  \\
& backbone &   2.31$\pm$1.85  & 1.44$\pm$1.78 \\
\midrule
\multirow{2}{*}{\makecell[c]{Attention\\ Type}} &self attention  & 1.61$\pm$0.64  & 1.33$\pm$1.75 \\
& Proposed & \textbf{1.27$\pm$0.46}  & \textbf{1.16$\pm$1.20} \\

\midrule
\multirow{5}{*}{\makecell[c]{Subject\\ Number}} 
& 20 & 2.10$\pm$2.27  & 1.88$\pm$2.71  \\
& 40 & 1.69$\pm$0.79  & 1.44$\pm$1.56  \\
& 60 & 1.56$\pm$0.90 & 1.38$\pm$1.73  \\
& 80  & 1.38$\pm$0.50 & 1.24$\pm$1.20 \\
& 100 & \textbf{1.27$\pm$0.46}  & \textbf{1.16$\pm$1.20} \\

\midrule
\multirow{3}{*}{\makecell[c]{PET\\Cloud\\ Size}} & $32^3$ & \textbf{1.27$\pm$0.46}  & \textbf{1.16$\pm$1.20}  \\
& $64^3$ & 1.45$\pm$0.78  & 1.37$\pm$1.75 \\
& $96^3$  & 1.59$\pm$0.60 & 1.49$\pm$1.85 \\
\bottomrule
\end{tabular*}
\end{center}
\label{all_ablation}
\end{table}

\subsection{DL-HMC++ Ablation Studies}

We conducted a series of ablation studies on the HRRT {$^{18}$F-FDG} dataset to evaluate individual components and select parameters that lead to the best motion estimation performance (\Cref{all_ablation}).

\subsubsection{Network Architecture} 
To demonstrate the effectiveness of the DL-HMC++ architecture, we compare 
\begin{inparaenum}[(i)]
    \item the proposed model architecture with self-gating and DNF; 
    \item the model without self-gating;
    \item the model without DNF; and
    \item the model without both self-gating and DNF.
\end{inparaenum}
DL-HMC++ without gating and DNF demonstrate the worse performance.
Removing the self-gating mechanism from the attention module degrades MC performance 0.25 mm in translation and 0.21{\textdegree} in rotation, which
demonstrates that our self-gating mechanism selectively distills the most relevant feature representation for motion tracking. 
Moreover, our results show that removing the DNF results in a performance drop of 22\% in translation and 13\% in rotation, which indicates that DNF plays a significant role in effectively aggregating information between the moving and reference branches to enhance the model's performance.

\subsubsection{Attention Type}
We experiment with different attention types:
\begin{inparaenum}[(i)]
    \item cross-attention, and
    \item self-attention.
\end{inparaenum}
Compared with the self-attention mechanism, which computes feature similarities within each input image individually, cross-attention concentrates feature learning on the head areas by computing the similarity between both the moving and reference clouds.
Quantitative evaluations demonstrate that our approach using cross-attention consistently outperforms self-attention in both translation and rotation. 
These results demonstrate that our approach boosts the model's MC performance by creating spatial correspondences between the moving and reference clouds.

\subsubsection{Training Set Size} 
We evaluate the impact of varying the number of subjects used for model training by evaluating performance using 20, 40, 60, 80, and 100 subjects.
As the number of subjects increases, we observe a corresponding enhancement in the performance of MC with a decrease in transformation error. 
DL-HMC++ achieves the best evaluation results on both translation and rotation using 100 subjects, demonstrating improvements of 39.5\% and 38.3\%, respectively, compared to the results when trained using 20 subjects. 
These results highlight the need for large training cohorts of PET studies when developing DL-based brain motion correction methods. 
\begin{figure}[t]
   \centering
   \includegraphics[width=0.85\columnwidth]{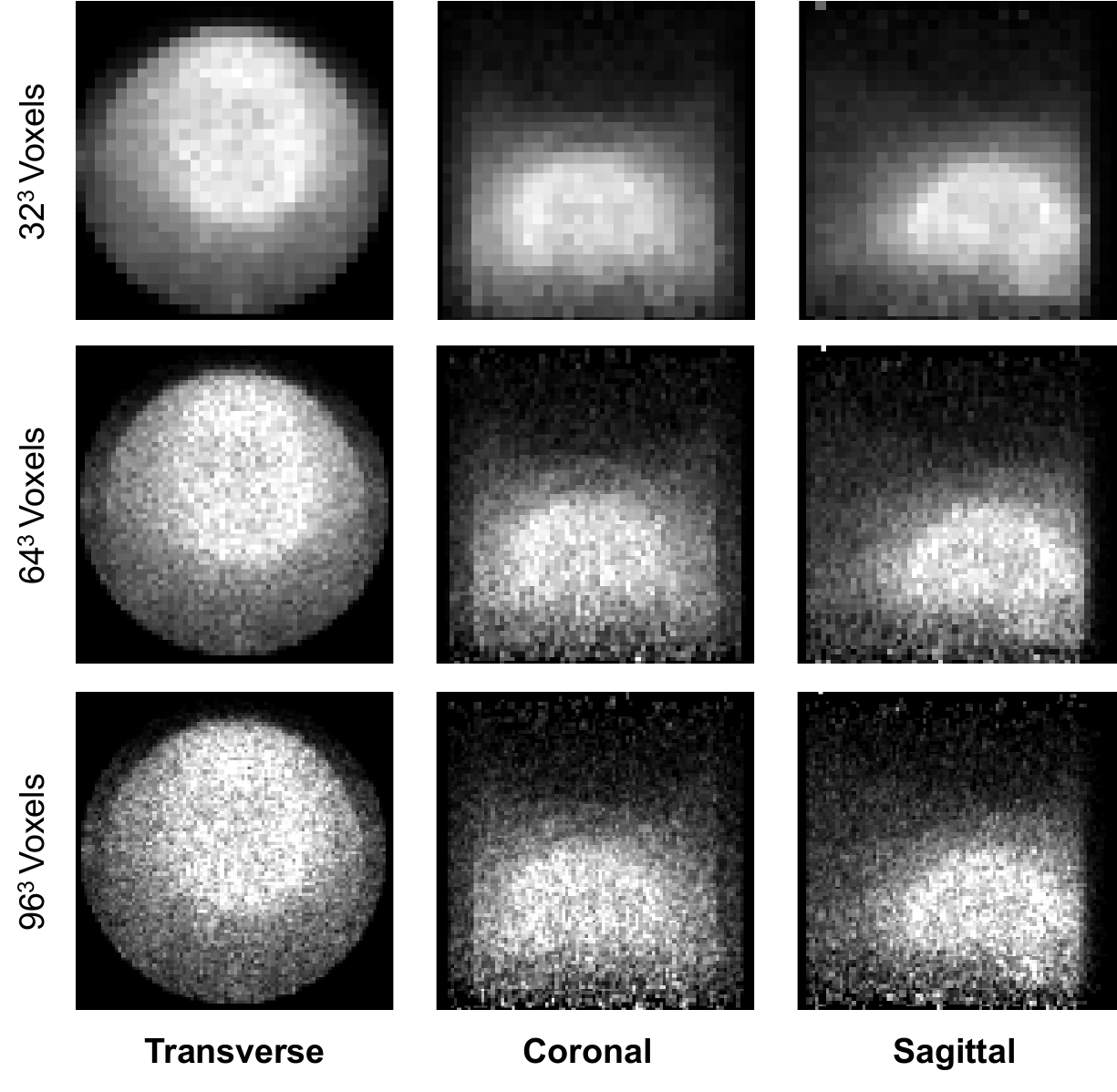}
   \caption{\textbf{3D PET Cloud Image (PCI) Dimensions.} Example one-second HRRT PET cloud images of different dimensions and resolutions: (top) 32$^3$ voxels; (middle) 64$^3$ voxels; and (bottom) 96$^3$ voxels.}
   \label{fig:pcis}
\end{figure}
\begin{table}[t]
\caption{
\small 
\textbf{Encoder Ablation Study.} Motion prediction of translation and rotation transformation {RMSE} on the HRRT $^{18}$F-FDG datasets. The encoder parameters, FLOPs, and inference time are also listed for comparison. Reported values are Mean$\pm$SD where appropriate.
}
\setlength\tabcolsep{3pt}

\begin{center}
\begin{tabular*}{\columnwidth}{l @{\extracolsep{\fill}} cccccc}
\toprule
   {Encoder} & \makecell[c]{Trans.\\(mm)}  & \makecell[c]{Rot.\\(deg)} & \makecell[c]{Parameters\\(M)} & \makecell[c]{FLOPs\\  ($\times10^9$)} & \makecell[c]{Inference\\ Time (ms)} \\
\midrule
    ResNet &1.62$\pm$0.83  & 1.37$\pm$1.88 &  14.61 & 4.6& 5.8\\
    U-Net & \textbf{1.27$\pm$0.46}  & \textbf{1.16$\pm$1.20}  &  \textbf{0.86} & \textbf{4.0} & \textbf{3.3} \\
\bottomrule
\end{tabular*}
\end{center}
\label{table: ablation_encoder}
\end{table}
\subsubsection{PET Cloud Image (PCI) Size}
We evaluate the performance of our model under various 3D PCI sizes: 32$^3$, 64$^3$, and 96$^3$.
As PCI size increases, there is a slight degradation in performance. 
Despite having lower spatial resolution, small PCI dimensions benefit from smooth images due to increased downsampling compared to larger PCIs (see \cref{fig:pcis}).
In contrast, the larger but noisier PCIs impair network training and fail to optimize motion correction performance.
\begin{figure*}[t]
    \includegraphics[width=\textwidth]{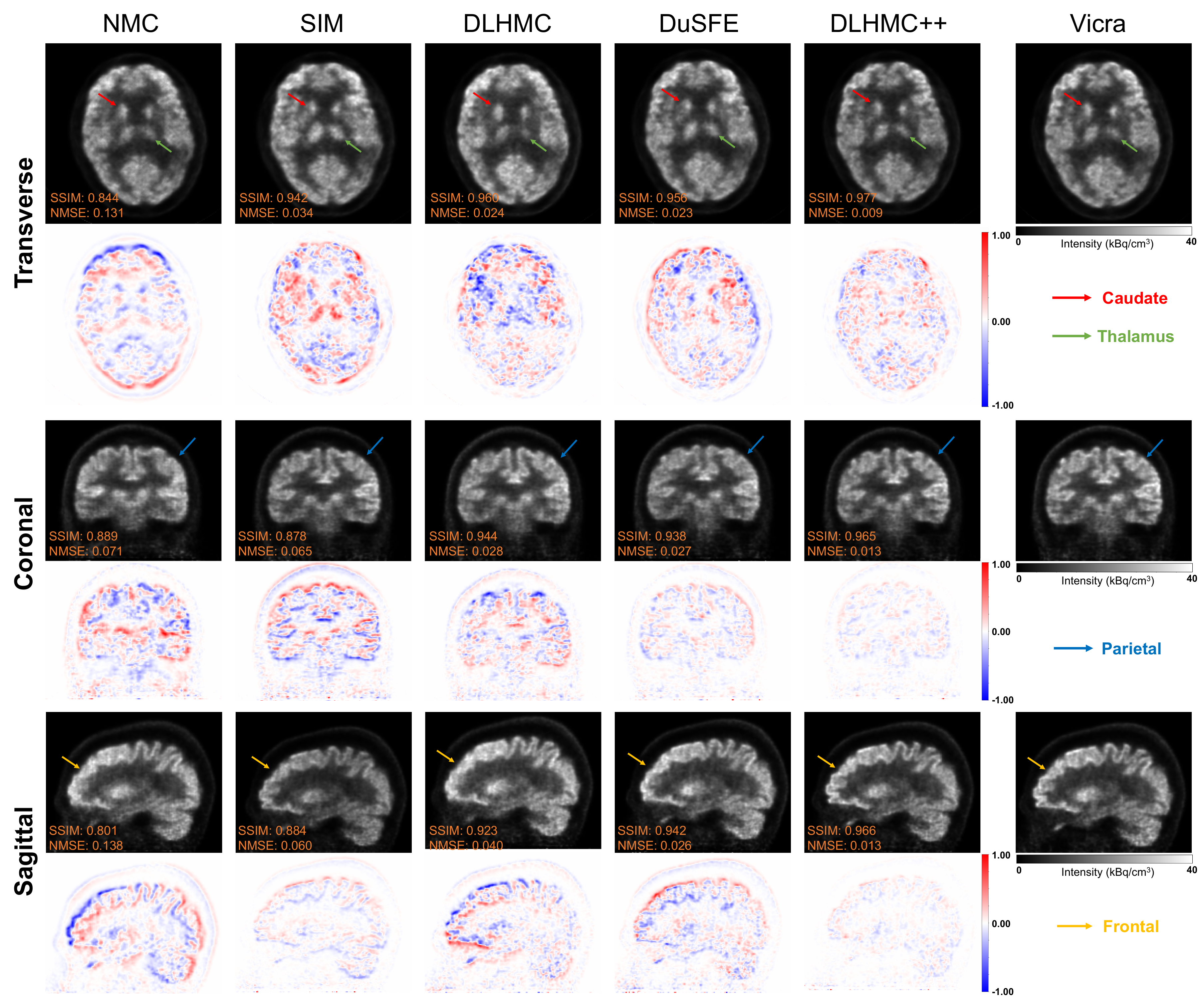}
    \vspace{-1.8\baselineskip}
    \caption{{\textbf{MOLAR Reconstruction comparison and error map between different MC methods for an HRRT $^{18}$F-FDG testing subject.} Arrows on reconstruction images indicate the specific brain ROI for visual comparison.} }
    \label{fig:recon_fdg}
\end{figure*}

\subsubsection{Network Encoder}
We further evaluate the choice of image encoder by comparing DL-HMC++'s U-Net encoder to DL-HMC's ResNet encoder, {removing the fully connected layer for a fair comparison.} As shown in \Cref{table: ablation_encoder}, we adopt the lightweight U-Net encoder instead of the ResNet encoder used in DL-HMC.
This change significantly reduces the number of {encoder parameters from 14.61M to 0.86M}, which enhances DL-HMC++ in terms of both training and inference efficiency.

\subsection{Motion-Corrected PET Image Reconstruction}
\subsubsection{Image Reconstruction Result}

\begin{figure*}[t]
    \centering
    \includegraphics[width=\textwidth]{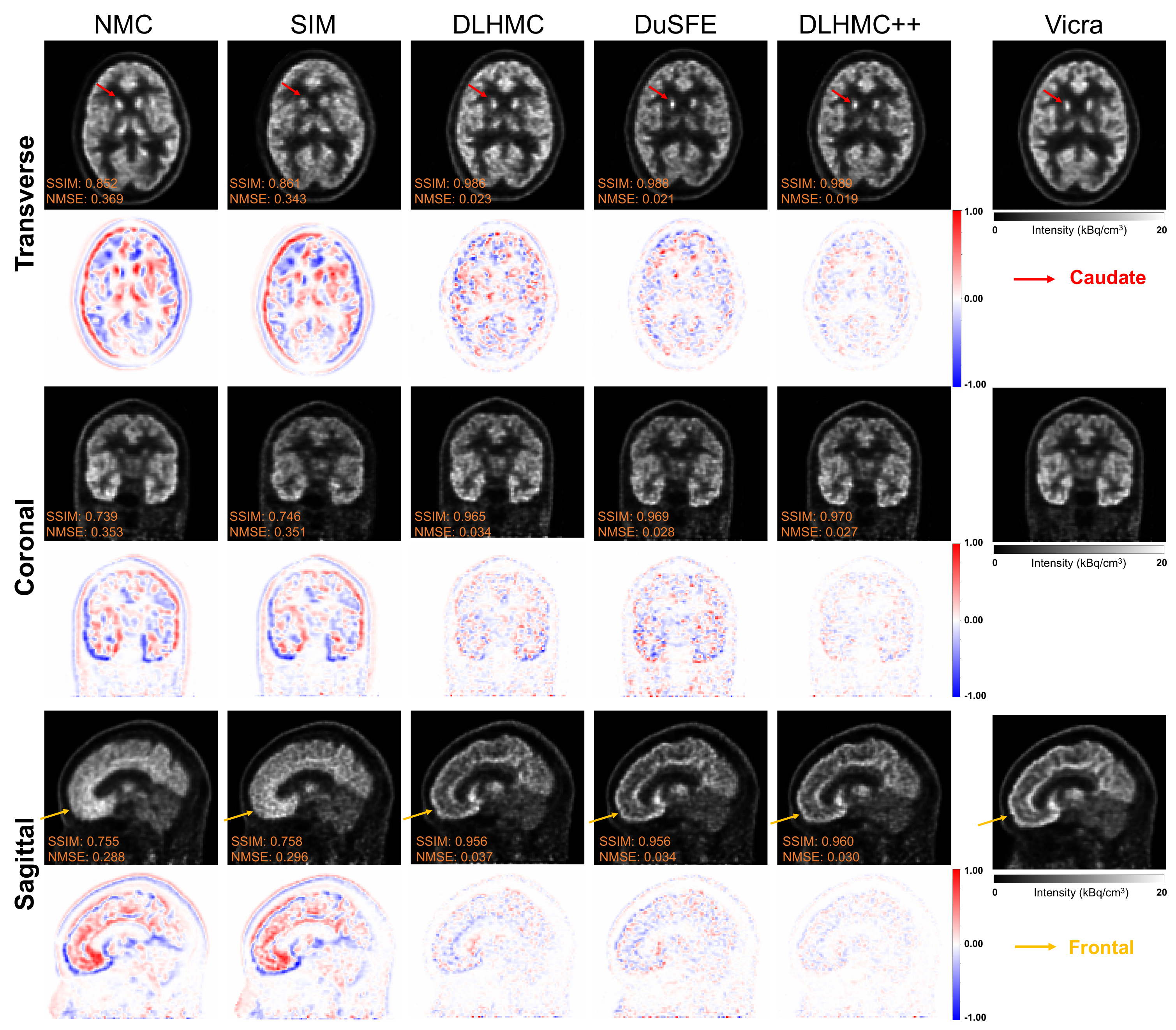}
    \vspace{-1.8\baselineskip}
    \caption{{\textbf{MOLAR Reconstruction comparison and error map between different MC methods for an mCT $^{18}$F-FPEB testing subject.} Arrows on reconstruction images indicate the specific brain ROI for visual comparison.} }
    \label{fig:recon_fpeb}
\end{figure*}

\Cref{fig:recon_fdg,fig:recon_fpeb} show MOLAR reconstruction images and {normalized} error maps with respect to Vicra gold-standard.
We randomly select one subject from the HRRT $^{18}$F-FDG testing set and one subject from the mCT $^{18}$F-FPEB testing set for visualization.
We compare reconstruction using DL-HMC++ to NMC, SIM, DuSFE, and DL-HMC with the Vicra gold-standard. 
Qualitatively, reconstruction using DL-HMC++ demonstrates the sharpest anatomical structure delineation and least deviation {(normalized error map)} from the Vicra gold-standard.{ Additionally, we compute the Structural Similarity Index (SSIM) and Normalized Mean Squared Error (NMSE) for each individual view to quantitatively assess image quality and reconstruction accuracy. } 

{In the HRRT $^{18}$F-FDG study, DL-HMC++-based reconstruction results clearly show the gyrus and sulcus on the entire cortex compared to NMC.
DL-HMC++ shows improved ROI anatomical structures such as the caudate and thalamus in the transverse view, as well as the parietal and frontal lobes in the coronal and sagittal views, respectively.}
{In addition, DL-HMC++ exhibits the highest SSIM, the lowest NMSE, and the smallest deviations from Vicra results compared to other methods, as indicated by the error maps.}

In the mCT $^{18}$F-FPEB study, NMC and SIM produce higher visual errors than the DL methods. {Notably, DLHMC++ achieves best quantification quality from SSIM and NMSE.}
{The transverse view} (\cref{fig:recon_fpeb}) { indicates that DL-HMC++ eliminates motion blurring for the caudate area, and the GM-WM interface can be delineated.}

\subsubsection{Brain ROI SUV Evaluation}
We average ROI SUV evaluation results across all 20 testing subjects in the HRRT $^{18}$F-FDG study and 4 testing subjects in the mCT $^{18}$F-FPEB study and compared percentage differences to the Vicra gold-standard (\cref{hrrtmct_roi}).
{Overall, DL-HMC++ outperforms all other methods, achieving the smallest mean SUV difference and the lowest standard deviation across both studies.
Compared to DL-HMC, DL-HMC++ demonstrates superior performance, with a 1.5\% improvement in mean SUV difference for $^{18}$F-FDG dataset and a 0.5\% improvement in $^{18}$F-FPEB dataset.}
{For $^{18}$F-FDG, the Wilcoxon signed-rank test indicates that the ROI SUV error of DL-HMC++ is significantly smaller than all other methods (p\textless0.05).}
For $^{18}$F-FPEB, DL-HMC++ and Vicra are nearly identical, with a 0.5\% average difference.
Notably, SIM performs worse than NMC, indicating that the intensity-based registration method with PCI input introduces false extra motion due to poor optimization. 

\subsubsection{MDE Evaluation Result}
\Cref{table:mde} presents the MDE metric result of all testing subjects in HRRT $^{18}$F-FDG and mCT $^{18}$F-FPEB studies. 
When evaluating anatomical brain ROI motion error, our results reveal a distinct advantage of DL methods over intensity-based methods with PCI input in terms of the MDE metric. 
In both studies, DL-HMC++ consistently demonstrates the smallest average MDE, underscoring the robustness and effectiveness of our proposed method.
Compared with DuSFE, DL-HMC++ not only achieves superior average MDE but also maintains lower standard deviation, indicating reduced variability of the proposed model. 
This reaffirms the superiority of DL-HMC++ in mitigating motion-related artifacts, rendering it a promising advancement in data-driven head motion estimation methods.

\subsection{Cross-tracer Generalization Performance}
\Cref{tab:cross_tracer} {summarizes the motion estimation RMSE results for two cross-tracer tasks using DL-HMC++. 
When compared to direct training on $^{18}$F-FDG, the cross-tracer experiment yields comparable results, with 0.23 mm higher for translation and 0.22\textdegree}{ higher for rotation. 
For $^{18}$F-FPEB, the cross-tracer results show 0.20 mm higher translation error and 0.15\textdegree}{ higher rotation error than directly training results, but still outperform all intensity-based registration methods and the DL-HMC method despite training with limited training data and different tracer characteristics.}

\begin{table*}[t]
\caption{
         \textbf{ROI evaluation result of different methods on HRRT and mCT.} The Absolute Difference Ratio (ADR) serves as the metric to quantify the discrepancy between different methods and Vicra gold-standard.
     } 
\begin{center}
\begin{tabular*}{\textwidth}{l @{\extracolsep{\fill}}|ccccc|ccccc}
\toprule
Dataset &  \multicolumn{5}{c}{HRRT $^{18}$F-FDG} \vline & \multicolumn{5}{c}{mCT $^{18}$F-FPEB}\\
\midrule
ROI\textbackslash ADR\% & NMC & SIM  & DL-HMC & DuSFE & DL-HMC++ & NMC & SIM & DL-HMC & DuSFE & DL-HMC++\\ 
\midrule
Amygdala & 6.8 & 6.8 & 2.1 & 1.9 & 1.7& 1.2 &	3.5 &	1.0 &	0.9 &	0.9\\
Caudate  & 13.8 &	11.8 & 5.6 & 2.4 & 2.0 & 4.6 &	10.2 &	2.2 &	0.6 &	0.6\\
Cerebellum Cortex & 13.8 & 11.8 & 5.6 & 2.4 &	2.0 & 0.4 &	0.7 &	0.3 &	0.2 &	0.2\\
Cerebellum WM & 5.6 &	5.5 &	1.3 &	0.7 &	0.6& 0.9 &	0.5 &	0.6 &	0.4 &	0.4\\
Cerebral WM & 4.3 &	3.5 &	2.0 &	1.1 &	1.1& 1.6 &	3.0 &	1.1 &	0.6 &	0.4\\
Frontal & 10.5 & 8.0 &	5.0 &	2.3 &	1.9& 2.9 &	5.2 &	1.5 &	0.6 &	0.7\\
Hippocampus & 7.9 & 6.6 & 2.0 &	0.9 &	0.9&2.6 &	3.3 &	1.9 &	1.2 &	0.7\\
Insula & 4.8 & 3.7 &	1.5 &	0.7 &	0.7& 1.8 &	4.1 &	0.5 &	0.5 &	0.3\\
Occipital & 8.6 &	8.6 &	3.2 &	1.7 &	1.5& 0.9 &	2.0 &	0.4 &	0.6 &	0.6\\
Pallidum & 4.5 &	3.4 &	1.4 &	1.0 &	1.0& 0.9 &	3.0 &	0.8 &	0.7 &	0.4\\
Parietal & 10.7 &	9.3 &	4.1 &	2.1 &	1.7& 1.9 &	3.4 &	0.9 &	0.6 &	0.5\\
Putamen & 8.7 &	6.9 &	3.3 &	1.0 &	1.1& 1.7 &	2.7 &	1.1 &	0.4 &	0.5\\
Temporal & 8.0 &	7.1 &	3.0 &	1.2 &	1.1& 1.3 &	3.1 &	0.9 &	0.4 &	0.4\\
Thalamus  & 9.7 &	7.7 &	2.6 &	1.0 &	0.9& 1.9 &	2.3 &	0.8 &	0.4 &	0.4\\
\midrule
Mean$\pm$SD & 7.9$\pm$2.7 &	6.8$\pm$2.3 &	2.7$\pm$1.3 &	1.4$\pm$0.6 &	\textbf{1.2$\pm$0.5} & 1.7$\pm$1.0&	3.3$\pm$2.2 &	1.0$\pm$0.5 &	0.6$\pm$0.2 &	\textbf{0.5$\pm$0.2}\\
\bottomrule
\end{tabular*}
\end{center}
\label{hrrtmct_roi}
\end{table*}

\begin{table}[t]
\caption{
\textbf{MDE metric for HRRT $^{18}$F-FDG and mCT $^{18}$F-FPEB studies.} Anatomical center of mass distance error metric compared to the gold-standard Vicra. Reported values in mm and are reported as Mean$\pm$SD.
} 
\begin{center}
\small
\begin{tabular*}{\columnwidth}{l @{\extracolsep{\fill}} cc}
\toprule
Method &  HRRT $^{18}$F-FDG &mCT $^{18}$F-FPEB\\
\midrule
NMC & 1.92$\pm$1.86 & 1.96$\pm$1.59 \\
SIM  & 1.86$\pm$0.54 & 1.59$\pm$0.53 \\
DL-HMC & 0.65$\pm$0.41 & 0.80$\pm$0.61 \\
DuSFE & 0.44$\pm$0.23 & 0.76$\pm$0.72 \\
DL-HMC++ & \textbf{0.39$\pm$0.11} & \textbf{0.65$\pm$0.66} \\ 
\bottomrule
\end{tabular*}
\end{center}
\label{table:mde}
\end{table}

\begin{table}[t]
\centering
\small
\caption{{\textbf{Cross-tracer Generalization RMSE Results.}}}
\label{tab:cross_tracer}

\begin{tabular*}{\columnwidth}{l@{\extracolsep{\fill}}cc}
\toprule
 Tasks & Trans. (mm) & Rot. (deg) \\
 \midrule 
 $^{18}$F-FDG NMC & 6.29$\pm$5.79 & 3.12$\pm$1.42 \\
$^{11}$C-UCB-J to $^{18}$F-FDG & 1.50$\pm$0.37 & 1.38$\pm$1.52  \\
DL-HMC++ on $^{18}$F-FDG & \textbf{1.27$\pm$0.46} & \textbf{1.16$\pm$1.20} \\
\midrule 
 $^{18}$F-FPEB NMC & 2.42$\pm$1.43 & 1.36$\pm$0.48 \\
$^{11}$C-LSN3172176 to $^{18}$F-FPEB &0.74$\pm$0.02 & 0.55$\pm$0.00 \\
DL-HMC++ on $^{18}$F-FPEB  & \textbf{0.54$\pm$0.00} & \textbf{0.40$\pm$0.00} \\
\bottomrule 
\end{tabular*}
\end{table}

\section{Discussion}
\label{sec:discussion}

DL-HMC++, a novel supervised deep learning model for PET head motion estimation with a cross-attention module, demonstrates effective motion estimation capabilities without the need for external hardware-based motion tracking (HMT) on testing subjects from two different scanners and four different tracers in a large cohort study.
Our evaluation on two different PET scanners, HRRT and mCT, {using four different tracers, $^{18}$F-FDG, $^{18}$F-FPEB, $^{11}$C-UCB-J, and $^{11}$C-LSN3172176}, shows that DL-HMC++ outperforms other benchmark SOTA methods, yielding motion tracking results similar to gold-standard Vicra HMT.
Qualitative and quantitative results demonstrate that the proposed method effectively eliminates motion blurring for head PET scans. 
In addition, we validate each contribution of our model design choices through comprehensive ablation studies.
By integrating the cross-attention mechanism, our model establishes spatial correspondences between the reference and moving PCIs, which enhances the ability of the model to track motion. 
Compared to the original DL-HMC implementation, the cross-attention mechanism guides the network to focus on motion-relevant information, diminishing the influence of irrelevant features. 
This process not only enhances the precision of the motion estimation but also improves robustness across the scan duration.
Remarkably, despite extremely {blurry} images (\cref{fig:pcis}), DL-HMC++ demonstrates anatomical motion errors of magnitude $\sim$1 mm (\cref{table:mde}) that are far below the input PCI {voxel size} of $\sim$10 mm$^3$ for both the HRRT and mCT studies.

\begin{table}[t]
\centering
\small
\caption{{\textbf{Comprehensive Ablation Study for IMR Method on the mCT $^{18}$F-FPEB dataset}}}
\label{tab:imrablation_results}
\begin{tabular*}{\columnwidth}{l@{\extracolsep{\fill}}cc}
\toprule
{Method} & Trans. (mm) & Rot. (deg) 
 \\ \midrule
IMR & 1.64$\pm$0.49 & 0.78$\pm$0.34 \\ 
w/o filter &  1.55$\pm$0.54  & 0.77$\pm$0.35 \\ 
w/o FRI & 4.30$\pm$6.31  & 1.43$\pm$0.46  \\ 
w/o dynamic reference & 1.53$\pm$0.40  & 0.76$\pm$0.34   \\ 
\bottomrule 
\end{tabular*}
\end{table}

The observed failures {and performance degradation for intensity-based registration methods on $^{11}$C dataset, e.g., the IMR result on $^{11}$C-LSN3172176 dataset (mean translation error 2.32 mm) compared to the $^{18}$F-FPEB dataset (mean translation error 1.38), are expected.} This is due to the intensity variations and noise in the dynamic input data, especially when comparing the appearance differences between the first reference time frame and the later frames. 
{To compare with SOTA data-driven motion tracking method, we implemented the IMR method following Spangler-Bickell's work}~\cite{spangler2021optimizing} {on the mCT dataset.
However, the motion estimation result reveals that all DL methods, especially DL-HMC++, outperform the IMR result.
In addition, we performed an ablation study for the IMR using 8 randomly selected subjects from the mCT $^{18}$F-FPEB dataset. Following optimization strategies in}~\cite{spangler2021optimizing}{, motion estimation performance without 6-mm Gaussian filtering, FRI input and dynamic reference frame were evaluated and the results are summarized in }\Cref{tab:imrablation_results}{. The IMR ablation result demonstrates that FRI is the primary contributor to the performance improvement of IMR, where filtering and dynamic reference frame did not affect the performance. 
Notably, compared with DL-HMC++, a significant limitation of applying IMR is the need to develop a fast reconstruction platform to support fast reconstruction frames, alongside the requirement for fine-tuning for different tracers.} 

In our studies, due to the patient's posture for the PET scan, movements {in the rotation along the Y-axis (vertical direction)} from all subjects were extremely small, making it challenging for the model to capture. One reason is that Y rotation is less frequent than X {(horizontal direction)} rotation and Z {(patient bed movement direction)} rotation, resulting in less variability in Y rotation for the model to learn. 
Additionally, Y rotation tends to have a small magnitude.
Both reasons make it more difficult for the model to capture Y rotation changes compared with translation changes. 
Two possible solutions can be used to alleviate this. One is to assign a higher weight to Y rotations in the loss function. The other is to perform data augmentation to increase the variability of Y rotations.

{We further compared the performance and computational efficiency of two deep learning methods with attention mechanism, DuSFE and DL-HMC++. As shown in}\protect~\Cref{figure: time comparison}{, the quantitative analysis demonstrates that DL-HMC++ achieves a 13.6\% improvement in translation and a 12.5\% improvement in rotation compared to DuSFE. Additionally, the lightweight architecture of our proposed framework substantially enhances our advantages. Specifically, DL-HMC++ shows a 37\% reduction in the number of parameters (2.2M vs. 3.5M), an 81\% decrease in computational cost (4.0G FLOPs vs. 21.3G FLOPs), and a 57\% faster inference time (3.30ms vs. 7.67ms). These enhancements highlight the efficiency of DL-HMC++ in terms of resource utilization and computational speed, making it a more viable option for potential real-world applications where computational resources and time are critical constraints. The consistent outperformance of DL-HMC++ across all datasets further underscores its robustness and reliability in motion estimation tasks.}

\begin{table}[t]
\caption{
\small 
{\textbf{Computational efficiency and Performance Comparison between DL-HMC++ and DuSFE.} 
Reported value of average translation and rotation
errors are the mean value of all datasets.
}}
\setlength\tabcolsep{3pt}

\begin{center}
\begin{tabular*}{\columnwidth}{l @{\extracolsep{\fill}} ccccccc}
\toprule
   {Method} & \makecell[c]{Parameters \\($\times$$10^6$)} & \makecell[c]{FLOPs \\ ($\times10^9$)} & \makecell[c]{Inference\\ Time (ms)} & \makecell[c]{ Memory\\ (GB)}  & \makecell[c]{Avg. \\Trans.}  & \makecell[c]{Avg. \\Rot.} \\
\midrule
    DuSFE & 3.5 & 21.3 &  7.67 & 30.3 & 1.18 & 0.96 \\
    DL-HMC++ & \textbf{2.2}  & \textbf{4.0}  &  \textbf{3.30} & \textbf{6.9} & \textbf{1.02} & \textbf{0.84}\\
\bottomrule
\end{tabular*}
\end{center}
\label{figure: time comparison}
\end{table}
{Through comprehensive experimentation across diverse tracer types and scanner cohorts} (\cref{tab:all_prediction}){, we identified performance improvements resulting from the removal of the time-conditioning module from the original DL-HMC architecture. Although this module was initially designed to enhance temporal information encoding, our findings indicate that it introduces redundancy: the sampling strategy and image data already provide sufficient temporal information. This redundancy leads the model to neglect spatial information, resulting in overfitting on the training data.}

{In the ablation study, we explored using different PCI sizes ranging from $32^3$ to $96^3$. The results indicate that increasing the voxel size of the cloud image led to a degradation in performance. A possible reason for this decline is the increase in noise levels and the corresponding decrease in the signal-to-noise ratio with larger dimensions. Our findings suggest that larger voxel sizes provide a more stable and robust signal representation, which is crucial for accurately detecting motion even under noisy conditions.}

{In the cross-tracer generalization experiment, we explored the possibility of using a pre-trained network on different tracer datasets. Due to the intrinsic characteristics of $^{11}$C, the PCIs are noisier and thus more challenging to train. By applying a network trained on such a difficult dataset to a dataset with more stable tracer dynamics at late time points (e.g., $^{18}$F), we demonstrated that DL-HMC++ exhibits generalizability across different tracers. 
Less intuitively, performing the cross-tracer experiment in the opposite manner, using a model pre-trained on $^{18}$F and applying to $^{11}$C at test time, suffered from model failure. 
Future studies are needed to study this cross-tracer phenomenon in detail.
Future work will also consider applying DL-HMC++ to other sites, using the pre-trained network with few-shot fine-tuning to ensure that the network adapts to site-specific variations.}

 {The motion estimation methods in our study estimate transformation metrics from different images generated from PET raw data. Theoretically, motion parameters can also be directly estimated from sinograms, and it is feasible to employ deep learning algorithms for this purpose. However, part of our dataset includes TOF information, which causes the sinogram size to be much larger than the image size. In the future, we will explore the possibility of applying DL-HMC++ to other domains, such as sinograms and COD traces.} 
 
The proposed DL-HMC++ method exhibits certain limitations. 
{Although DL-HMC++ achieves comparable motion tracking results with short half-life $^{11}$C tracers, it exhibits a notable constraint in its inability to effectively detect motion during periods of rapid tracer dynamic changes, such as the first 10 minutes post-injection.}
Moreover, Vicra failure and inaccuracy may have a negative effect on the proposed supervised model.
{In the future, we aim to develop a generalized model for various tracers and scanners, including an ultra-high performance human brain PET/CT scanner}~\cite{li2024performance}{, which has a spatial resolution of less than 2.0 mm and is more sensitive to motion effects.} 
We will also investigate the feasibility of applying semi-supervised learning and unsupervised learning for PET head motion estimation~\cite{zeng2023teacher}.
\section{Conclusion}

In this paper, we proposed DL-HMC++ to predict head motion directly from {PET raw data to achieve robust data-driven head motion estimation}. 
DL-HMC++ incorporates a cross-attention mechanism to compute the correlation between two one-second PET cloud images. Cross-attention boosts the model's ability to track the motion by establishing spatial correspondence between the two images to be registered and focuses network learning on the most important regions of the image for head motion. 
We validated DL-HMC++ in a large cohort PET study with {4 different tracers on more than 280 subjects}, and the results demonstrated significant motion estimation performance improvements both qualitatively and quantitatively compared to {SOTA data-driven head motion estimation methods.
Extensive evaluation and ablation studies demonstrate the superior performance and feasibility of our proposed DL-HMC++ to address head motion estimation for PET without the need for hardware-based motion tracking.
Furthermore, the cross-tracer generalization experiment highlights the potential of the proposed network to effectively generalize across various tracers.}

\bibliographystyle{IEEEtran}

\begin{thebibliography}{10}
\providecommand{\url}[1]{#1}
\csname url@samestyle\endcsname
\providecommand{\newblock}{\relax}
\providecommand{\bibinfo}[2]{#2}
\providecommand{\BIBentrySTDinterwordspacing}{\spaceskip=0pt\relax}
\providecommand{\BIBentryALTinterwordstretchfactor}{4}
\providecommand{\BIBentryALTinterwordspacing}{\spaceskip=\fontdimen2\font plus
\BIBentryALTinterwordstretchfactor\fontdimen3\font minus \fontdimen4\font\relax}
\providecommand{\BIBforeignlanguage}[2]{{%
\expandafter\ifx\csname l@#1\endcsname\relax
\typeout{** WARNING: IEEEtran.bst: No hyphenation pattern has been}%
\typeout{** loaded for the language `#1'. Using the pattern for}%
\typeout{** the default language instead.}%
\else
\language=\csname l@#1\endcsname
\fi
#2}}
\providecommand{\BIBdecl}{\relax}
\BIBdecl

\bibitem{nabulsi2016synthesis}
N.~B. Nabulsi, J.~Mercier, D.~Holden, S.~Carr{\'e}, S.~Najafzadeh, M.-C. Vandergeten, S.-f. Lin, A.~Deo, N.~Price, M.~Wood \emph{et~al.}, ``Synthesis and preclinical evaluation of 11c-ucb-j as a pet tracer for imaging the synaptic vesicle glycoprotein 2a in the brain,'' \emph{Journal of Nuclear Medicine}, vol.~57, no.~5, pp. 777--784, 2016.

\bibitem{scholl2016pet}
M.~Sch{\"o}ll, S.~N. Lockhart, D.~R. Schonhaut, J.~P. O’Neil, M.~Janabi, R.~Ossenkoppele, S.~L. Baker, J.~W. Vogel, J.~Faria, H.~D. Schwimmer \emph{et~al.}, ``Pet imaging of tau deposition in the aging human brain,'' \emph{Neuron}, vol.~89, no.~5, pp. 971--982, 2016.

\bibitem{toyonaga2019vivo}
T.~Toyonaga, L.~M. Smith, S.~J. Finnema, J.-D. Gallezot, M.~Naganawa, J.~Bini, T.~Mulnix, Z.~Cai, J.~Ropchan, Y.~Huang \emph{et~al.}, ``In vivo synaptic density imaging with 11c-ucb-j detects treatment effects of saracatinib in a mouse model of alzheimer disease,'' \emph{Journal of Nuclear Medicine}, vol.~60, no.~12, pp. 1780--1786, 2019.

\bibitem{sarikaya2015pet}
I.~Sarikaya, ``Pet studies in epilepsy,'' \emph{American journal of nuclear medicine and molecular imaging}, vol.~5, no.~5, p. 416, 2015.

\bibitem{keller2012methods}
S.~H. Keller, M.~Sibomana, O.~V. Olesen, C.~Svarer, S.~Holm, F.~L. Andersen, and L.~H{\o}jgaard, ``Methods for motion correction evaluation using 18f-fdg human brain scans on a high-resolution pet scanner,'' \emph{Journal of Nuclear Medicine}, vol.~53, no.~3, pp. 495--504, 2012.

\bibitem{beyer2005use}
T.~Beyer, L.~Tellmann, I.~Nickel, and U.~Pietrzyk, ``On the use of positioning aids to reduce misregistration in the head and neck in whole-body pet/ct studies,'' \emph{Journal of Nuclear Medicine}, vol.~46, no.~4, pp. 596--602, 2005.

\bibitem{rahmim2007strategies}
A.~Rahmim, O.~Rousset, and H.~Zaidi, ``Strategies for motion tracking and correction in pet,'' \emph{PET clinics}, vol.~2, no.~2, pp. 251--266, 2007.

\bibitem{picard1997motion}
Y.~Picard and C.~J. Thompson, ``Motion correction of pet images using multiple acquisition frames,'' \emph{IEEE transactions on medical imaging}, vol.~16, no.~2, pp. 137--144, 1997.

\bibitem{jin2013evaluation}
X.~Jin, T.~Mulnix, J.-D. Gallezot, and R.~E. Carson, ``Evaluation of motion correction methods in human brain pet imaging—a simulation study based on human motion data,'' \emph{Medical physics}, vol.~40, no.~10, p. 102503, 2013.

\bibitem{montgomery2006correction}
A.~J. Montgomery, K.~Thielemans, M.~A. Mehta, F.~Turkheimer, S.~Mustafovic, and P.~M. Grasby, ``Correction of head movement on pet studies: comparison of methods,'' \emph{Journal of Nuclear Medicine}, vol.~47, no.~12, pp. 1936--1944, 2006.

\bibitem{sun2022objective}
C.~Sun, E.~M. Revilla, J.~Zhang, K.~Fontaine, T.~Toyonaga, J.-D. Gallezot, T.~Mulnix, J.~A. Onofrey, R.~E. Carson, and Y.~Lu, ``An objective evaluation method for head motion estimation in pet—motion corrected centroid-of-distribution,'' \emph{NeuroImage}, vol. 264, p. 119678, 2022.

\bibitem{iwao2022marker}
Y.~Iwao, G.~Akamatsu, H.~Tashima, M.~Takahashi, and T.~Yamaya, ``Marker-less and calibration-less motion correction method for brain pet,'' \emph{Radiological Physics and Technology}, vol.~15, no.~2, pp. 125--134, 2022.

\bibitem{slipsager2019markerless}
J.~M. Slipsager, A.~H. Ellegaard, S.~L. Glimberg, R.~R. Paulsen, M.~D. Tisdall, P.~Wighton, A.~Van Der~Kouwe, L.~Marner, O.~M. Henriksen, I.~Law \emph{et~al.}, ``Markerless motion tracking and correction for pet, mri, and simultaneous pet/mri,'' \emph{PloS one}, vol.~14, no.~4, p. e0215524, 2019.

\bibitem{zeng2023markerless}
T.~Zeng, Y.~Lu, W.~Jiang, J.~Zheng, J.~Zhang, P.~Gravel, Q.~Wan, K.~Fontaine, T.~Mulnix, Y.~Jiang \emph{et~al.}, ``Markerless head motion tracking and event-by-event correction in brain pet,'' \emph{Physics in Medicine \& Biology}, vol.~68, no.~24, p. 245019, 2023.

\bibitem{zhang2024data}
J.~Zhang, C.~Sun, T.~Volpi, T.~Zeng, K.~Fontaine, Y.~Du, T.~Toyonaga, J.~A. Onofrey, Y.~Lu, and R.~E. Carson, ``Data-driven non-rigid motion detection and correction for neuroexplorer,'' in \emph{2024 IEEE Nuclear Science Symposium (NSS), Medical Imaging Conference (MIC) and Room Temperature Semiconductor Detector Conference (RTSD)}.\hskip 1em plus 0.5em minus 0.4em\relax IEEE, 2024, pp. 1--2.

\bibitem{spangler2022evaluation}
M.~G. Spangler-Bickell, S.~A. Hurley, A.~Pirasteh, S.~B. Perlman, T.~Deller, and A.~B. McMillan, ``Evaluation of data-driven rigid motion correction in clinical brain pet imaging,'' \emph{Journal of Nuclear Medicine}, vol.~63, no.~10, pp. 1604--1610, 2022.

\bibitem{revilla2022adaptive}
E.~M. Revilla, J.-D. Gallezot, M.~Naganawa, T.~Toyonaga, K.~Fontaine, T.~Mulnix, J.~A. Onofrey, R.~E. Carson, and Y.~Lu, ``Adaptive data-driven motion detection and optimized correction for brain pet,'' \emph{Neuroimage}, vol. 252, p. 119031, 2022.

\bibitem{lecun2015deep}
Y.~LeCun, Y.~Bengio, and G.~Hinton, ``Deep learning,'' \emph{nature}, vol. 521, no. 7553, pp. 436--444, 2015.

\bibitem{fu2020deep}
Y.~Fu, Y.~Lei, T.~Wang, W.~J. Curran, T.~Liu, and X.~Yang, ``Deep learning in medical image registration: a review,'' \emph{Physics in Medicine \& Biology}, vol.~65, no.~20, p. 20TR01, 2020.

\bibitem{salehi2018real}
S.~S.~M. Salehi, S.~Khan, D.~Erdogmus, and A.~Gholipour, ``Real-time deep pose estimation with geodesic loss for image-to-template rigid registration,'' \emph{IEEE transactions on medical imaging}, vol.~38, no.~2, pp. 470--481, 2018.

\bibitem{de2019deep}
B.~D. De~Vos, F.~F. Berendsen, M.~A. Viergever, H.~Sokooti, M.~Staring, and I.~I{\v{s}}gum, ``A deep learning framework for unsupervised affine and deformable image registration,'' \emph{Medical image analysis}, vol.~52, pp. 128--143, 2019.

\bibitem{zeng2022supervised}
T.~Zeng, J.~Zhang, E.~Revilla, E.~V. Lieffrig, X.~Fang, Y.~Lu, and J.~A. Onofrey, ``Supervised deep learning for head motion correction in pet,'' in \emph{International Conference on Medical Image Computing and Computer-Assisted Intervention}.\hskip 1em plus 0.5em minus 0.4em\relax Springer, 2022, pp. 194--203.

\bibitem{sundar2021conditional}
L.~K.~S. Sundar, D.~Iommi, O.~Muzik, Z.~Chalampalakis, E.-M. Klebermass, M.~Hienert, L.~Rischka, R.~Lanzenberger, A.~Hahn, E.~Pataraia \emph{et~al.}, ``Conditional generative adversarial networks aided motion correction of dynamic 18f-fdg pet brain studies,'' \emph{Journal of Nuclear Medicine}, vol.~62, no.~6, pp. 871--879, 2021.

\bibitem{Lieffrig2023-nw}
E.~V. Lieffrig, T.~Zeng, J.~Zhang, K.~Fontaine, X.~Fang, E.~Revilla, Y.~Lu, and J.~A. Onofrey, ``Multi-task deep learning and uncertainty estimation for pet head motion correction,'' in \emph{2023 IEEE 20th International Symposium on Biomedical Imaging (ISBI)}.\hskip 1em plus 0.5em minus 0.4em\relax IEEE, 2023, pp. 1--5.

\bibitem{reimers2023deep}
E.~Reimers, J.-C. Cheng, and V.~Sossi, ``Deep-learning-aided intraframe motion correction for low-count dynamic brain pet,'' \emph{IEEE Transactions on Radiation and Plasma Medical Sciences}, vol.~8, no.~1, pp. 53--63, 2023.

\bibitem{Cai2023-ic}
Z.~Cai, T.~Zeng, E.~V. Lieffrig, J.~Zhang, F.~Chen, T.~Toyonaga, C.~You, J.~Xin, N.~Zheng, Y.~Lu \emph{et~al.}, ``Cross-attention for improved motion correction in brain pet,'' in \emph{International Workshop on Machine Learning in Clinical Neuroimaging}.\hskip 1em plus 0.5em minus 0.4em\relax Springer, 2023, pp. 34--45.

\bibitem{Ahn2023-dk}
S.~S. Ahn, K.~Ta, S.~L. Thorn, J.~A. Onofrey, I.~H. Melvinsdottir, S.~Lee, J.~Langdon, A.~J. Sinusas, and J.~S. Duncan, ``Co-attention spatial transformer network for unsupervised motion tracking and cardiac strain analysis in 3d echocardiography,'' \emph{Medical image analysis}, vol.~84, p. 102711, 2023.

\bibitem{chen2022dual}
X.~Chen, B.~Zhou, H.~Xie, X.~Guo, J.~Zhang, A.~J. Sinusas, J.~A. Onofrey, and C.~Liu, ``Dual-branch squeeze-fusion-excitation module for cross-modality registration of cardiac spect and ct,'' in \emph{Medical Image Computing and Computer Assisted Intervention--MICCAI 2022: 25th International Conference, Singapore, September 18--22, 2022, Proceedings, Part VI}.\hskip 1em plus 0.5em minus 0.4em\relax Springer, 2022, pp. 46--55.

\bibitem{ronneberger2015u}
O.~Ronneberger, P.~Fischer, and T.~Brox, ``U-net: Convolutional networks for biomedical image segmentation,'' in \emph{Medical Image Computing and Computer-Assisted Intervention--MICCAI 2015: 18th International Conference, Munich, Germany, October 5-9, 2015, Proceedings, Part III 18}.\hskip 1em plus 0.5em minus 0.4em\relax Springer, 2015, pp. 234--241.

\bibitem{mu2023learning}
P.~Mu, G.~Wu, J.~Liu, Y.~Zhang, X.~Fan, and R.~Liu, ``Learning to search a lightweight generalized network for medical image fusion,'' \emph{IEEE Transactions on Circuits and Systems for Video Technology}, 2023.

\bibitem{chen2021comparison}
M.-K. Chen, A.~P. Mecca, M.~Naganawa, J.-D. Gallezot, T.~Toyonaga, J.~Mondal, S.~J. Finnema, S.-f. Lin, R.~S. O’Dell, J.~W. McDonald \emph{et~al.}, ``Comparison of [11c] ucb-j and [18f] fdg pet in alzheimer's disease: a tracer kinetic modeling study,'' \emph{Journal of Cerebral Blood Flow \& Metabolism}, vol.~41, no.~9, pp. 2395--2409, 2021.

\bibitem{lim2014preparation}
K.~Lim, D.~Labaree, S.~Li, and Y.~Huang, ``Preparation of the metabotropic glutamate receptor 5 (mglur5) pet tracer [18f] fpeb for human use: an automated radiosynthesis and a novel one-pot synthesis of its radiolabeling precursor,'' \emph{Applied Radiation and Isotopes}, vol.~94, pp. 349--354, 2014.

\bibitem{naganawa2021first}
M.~Naganawa, N.~Nabulsi, S.~Henry, D.~Matuskey, S.-F. Lin, L.~Slieker, A.~J. Schwarz, N.~Kant, C.~Jesudason, K.~Ruley \emph{et~al.}, ``First-in-human assessment of 11c-lsn3172176, an m1 muscarinic acetylcholine receptor pet radiotracer,'' \emph{Journal of Nuclear Medicine}, vol.~62, no.~4, pp. 553--560, 2021.

\bibitem{fischl2012freesurfer}
B.~Fischl, ``Freesurfer,'' \emph{Neuroimage}, vol.~62, no.~2, pp. 774--781, 2012.

\bibitem{papademetris2006bioimage}
X.~Papademetris, M.~P. Jackowski, N.~Rajeevan, M.~DiStasio, H.~Okuda, R.~T. Constable, and L.~H. Staib, ``Bioimage suite: An integrated medical image analysis suite: An update,'' \emph{The insight journal}, p. 209, 2006.

\bibitem{marstal2016simpleelastix}
K.~Marstal, F.~Berendsen, M.~Staring, and S.~Klein, ``Simpleelastix: A user-friendly, multi-lingual library for medical image registration,'' in \emph{Proceedings of the IEEE conference on computer vision and pattern recognition workshops}, 2016, pp. 134--142.

\bibitem{muthukumaran2017medical}
D.~Muthukumaran and M.~Sivakumar, ``Medical image registration: a matlab based approach,'' \emph{Int. J. Sci. Res. Comput. Sci. Eng. Inf. Technol}, vol.~2, no.~1, pp. 29--34, 2017.

\bibitem{lieffrig2022multi}
E.~V. Lieffrig, T.~Zeng, J.~Zhang, X.~Fang, E.~Revilla, Y.~Lu, and J.~A. Onofrey, ``Multi-tracer deep learning for pet head motion correction,'' in \emph{2022 IEEE Nuclear Science Symposium and Medical Imaging Conference (NSS/MIC)}.\hskip 1em plus 0.5em minus 0.4em\relax IEEE, 2022, pp. 1--4.

\bibitem{chen2023dusfe}
X.~Chen, B.~Zhou, H.~Xie, X.~Guo, J.~Zhang, J.~S. Duncan, E.~J. Miller, A.~J. Sinusas, J.~A. Onofrey, and C.~Liu, ``Dusfe: Dual-channel squeeze-fusion-excitation co-attention for cross-modality registration of cardiac spect and ct,'' \emph{Medical Image Analysis}, vol.~88, p. 102840, 2023.

\bibitem{spangler2021optimizing}
M.~G. Spangler-Bickell, S.~A. Hurley, T.~W. Deller, F.~Jansen, V.~Bettinardi, M.~Carlson, M.~Zeineh, G.~Zaharchuk, and A.~B. McMillan, ``Optimizing the frame duration for data-driven rigid motion estimation in brain pet imaging,'' \emph{Medical physics}, vol.~48, no.~6, pp. 3031--3041, 2021.

\bibitem{zeng2023fast}
T.~Zeng, J.~Zhang, E.~V. Lieffrig, Z.~Cai, F.~Chen, C.~You, M.~Naganawa, Y.~Lu, and J.~A. Onofrey, ``{Fast Reconstruction for Deep Learning PET Head Motion Correction},'' in \emph{International Conference on Medical Image Computing and Computer-Assisted Intervention}.\hskip 1em plus 0.5em minus 0.4em\relax Springer, 2023, pp. 710--719.

\bibitem{jin2014evaluation}
X.~Jin, T.~Mulnix, C.~M. Sandiego, and R.~E. Carson, ``Evaluation of frame-based and event-by-event motion-correction methods for awake monkey brain pet imaging,'' \emph{Journal of Nuclear Medicine}, vol.~55, no.~2, pp. 287--293, 2014.

\bibitem{selvaraju2017grad}
R.~R. Selvaraju, M.~Cogswell, A.~Das, R.~Vedantam, D.~Parikh, and D.~Batra, ``Grad-cam: Visual explanations from deep networks via gradient-based localization,'' in \emph{Proceedings of the IEEE international conference on computer vision}, 2017, pp. 618--626.

\bibitem{li2024performance}
H.~Li, R.~D. Badawi, S.~R. Cherry, K.~Fontaine, L.~He, S.~Henry, A.~T. Hillmer, L.~Hu \emph{et~al.}, ``Performance characteristics of the neuroexplorer, a next-generation human brain pet/ct imager,'' \emph{Journal of Nuclear Medicine}, vol.~65, no.~8, pp. 1320--1326, 2024.

\bibitem{zeng2023teacher}
T.~Zeng, C.~You, Z.~Cai, E.~V. Lieffrig, J.~Zhang, F.~Chen, Y.~Lu, and J.~A. Onofrey, ``Teacher’s pet: Semi-supervised deep learning for pet head motion correction,'' in \emph{2023 IEEE Nuclear Science Symposium (NSS), Medical Imaging Conference (MIC) and Room Temperature Semiconductor Detector Conference (RTSD)}.\hskip 1em plus 0.5em minus 0.4em\relax IEEE, 2023, pp. 1--1.

\end{thebibliography}

\clearpage

\setcounter{page}{1}
\setcounter{section}{0}
\setcounter{figure}{0}
\setcounter{table}{0}
\renewcommand{\thepage}{S\arabic{page}} 
\renewcommand{\thesection}{S\arabic{section}}
\renewcommand{\thefigure}{S\arabic{figure}}%
\renewcommand{\thetable}{S\arabic{table}}%

\end{document}